\documentclass{article}
\usepackage{iclr2027_conference,times}

\usepackage[utf8]{inputenc} 
\usepackage[T1]{fontenc}    
\usepackage{hyperref}       
\usepackage{url}            
\usepackage{booktabs}       
\usepackage{amsfonts}       
\usepackage{nicefrac}       
\usepackage{microtype}      
\usepackage{xcolor}         

\usepackage[table]{xcolor}
\usepackage{subcaption}
\usepackage{graphicx} 
\usepackage{wrapfig}
\usepackage{arydshln}
\usepackage{fvextra}
\usepackage{amsmath}
\usepackage{bbm}
\usepackage{soul}
\usepackage[most]{tcolorbox}

\definecolor{thoughtcolor}{RGB}{35, 92, 160}
\definecolor{reflectioncolor}{RGB}{35, 92, 160}
\definecolor{actioncolor}{RGB}{190, 75, 45}
\definecolor{observationcolor}{RGB}{45, 125, 80}
\definecolor{taskcolor}{RGB}{125, 70, 145}
\definecolor{reasoningbg}{RGB}{255, 248, 220}
\sethlcolor{reasoningbg}

\newcommand{\hlreason}[1]{\hl{#1}}
\newcommand{\thoughtlabel}{%
    \textcolor{thoughtcolor}{\textbf{Thought: }}%
}
\newcommand{\reflectionlabel}{%
    \textcolor{reflectioncolor}{\textbf{Reflection: }}%
}
\newcommand{\actionlabel}{%
    \textcolor{actioncolor}{\textbf{Action: }}%
}
\newcommand{\observationlabel}{%
    \textcolor{observationcolor}{\textbf{Observation: }}%
}
\newcommand{\tasklabel}{%
    \textcolor{taskcolor}{\textbf{Your task is to: }}%
}
\newcommand{\promptgap}{\par\vspace{0.6em}}

\newcommand{\promptheading}[1]{%
    \vspace{0.75em}
    \noindent
    \colorbox{black!8}{%
        \parbox{\dimexpr\linewidth-2\fboxsep\relax}{%
            \textbf{\small #1}%
        }%
    }%
    \par\vspace{0.25em}
}

\newtcolorbox{PromptBlock}{
    colback=white,
    colframe=black,
    boxrule=0.5pt,
    arc=0pt,
    left=2mm,
    right=2mm,
    top=2mm,
    bottom=2mm,
    breakable,
    fontupper=\ttfamily\footnotesize
}

\title{Towards a Belief-Based World Model for LLM Agents}

\author{%
  Shubham Kumar \\
  UIUC\\      
  \And
  Harshit Kumar \\
  IBM \\  
  \And
  Narendra Ahuja \\
  UIUC \\    
  \And
  Saurabh Jha \\
  IBM \\
}

\iclrfinalcopy
\begin{document}

\maketitle
\lhead{}

\begin{abstract}
  Large language models (LLMs) are being used as policies for autonomous decision-making and planning in many domains. Despite their strong reasoning capabilities, LLMs struggle with long-horizon tasks, especially under partial observability. World models are a promising way to enhance policy performance, both during training and inference. During inference, agents currently use world models to simulate the consequences of candidate actions before choosing an action, which can improve decision-making. However, we argue that simulation alone is an incomplete interface for decision-making under partial observability: simulation does not adequately capture uncertainty about the current state, which agents may need for accurate decision-making. We address this limitation with Belief-Based World Models (BB-WMs), which maintain a \textit{belief} that LLMs can query to access information on what is known and uncertain about the current state. Before developing methods to learn accurate BB-WMs, this paper focuses on a more fundamental question: does exposing a world model's belief directly to an LLM policy improve decision-making? Our results show that giving LLM agents access to beliefs improves task performance under partial observability, while remaining complementary to existing simulation-based world models. Code: \href{https://github.com/skumar-ml/belief-world-models}{https://github.com/skumar-ml/belief-world-models}.
\end{abstract}

\section{Introduction}

Consider an autonomous system that plans and executes a series of actions to achieve some goal. It's hypothesized that creating models of the world (or \textit{world models}) can improve the learning of the system's decision-making function, or \textit{policy}. Instead of needing costly real-world interactions, a policy can be trained on diverse, imagined future trajectories from a world model \citep{hafner2020dreamer, haAndSchmidhuber2018worldmodels}. World modeling training objectives can also be used alongside reinforcement learning (RL) to improve the policy itself \citep{Yu2026ReinforcementWM, RLwithSelfPredictiveReps, CodeWorldModel}. Recent world modeling efforts have primarily gone towards improving the policy at training time \citep{GenieGoogle, Agarwal2025CosmosWF, Ren2025CosmosDriveDreamsSS}. Once trained, the policy does not use the world model within a control loop (i.e., \textit{model-free control}). While this is a promising direction, it is unclear how well such policies generalize to novel situations at test time. To help with this, the policy may also have access to a world model at test-time (i.e., \textit{model-based control}), where foresight from the world model enables test-time reasoning (e.g., planning) in difficult or novel situations. This direction has received comparatively less attention and is the focus of this work.

\begin{figure}[tbp]
    \centering
    \setlength{\tabcolsep}{3pt}
    \renewcommand{\arraystretch}{1.1}

    \begin{tabular}{c : cc : c}
        \multicolumn{1}{c:}{\small\textbf{Setting}} &
        \multicolumn{2}{c:}{\small\textbf{Simulation Interface}} &
        \small\textbf{Belief} \\[2pt]

        \includegraphics[width=0.225\textwidth,height=4cm,keepaspectratio]{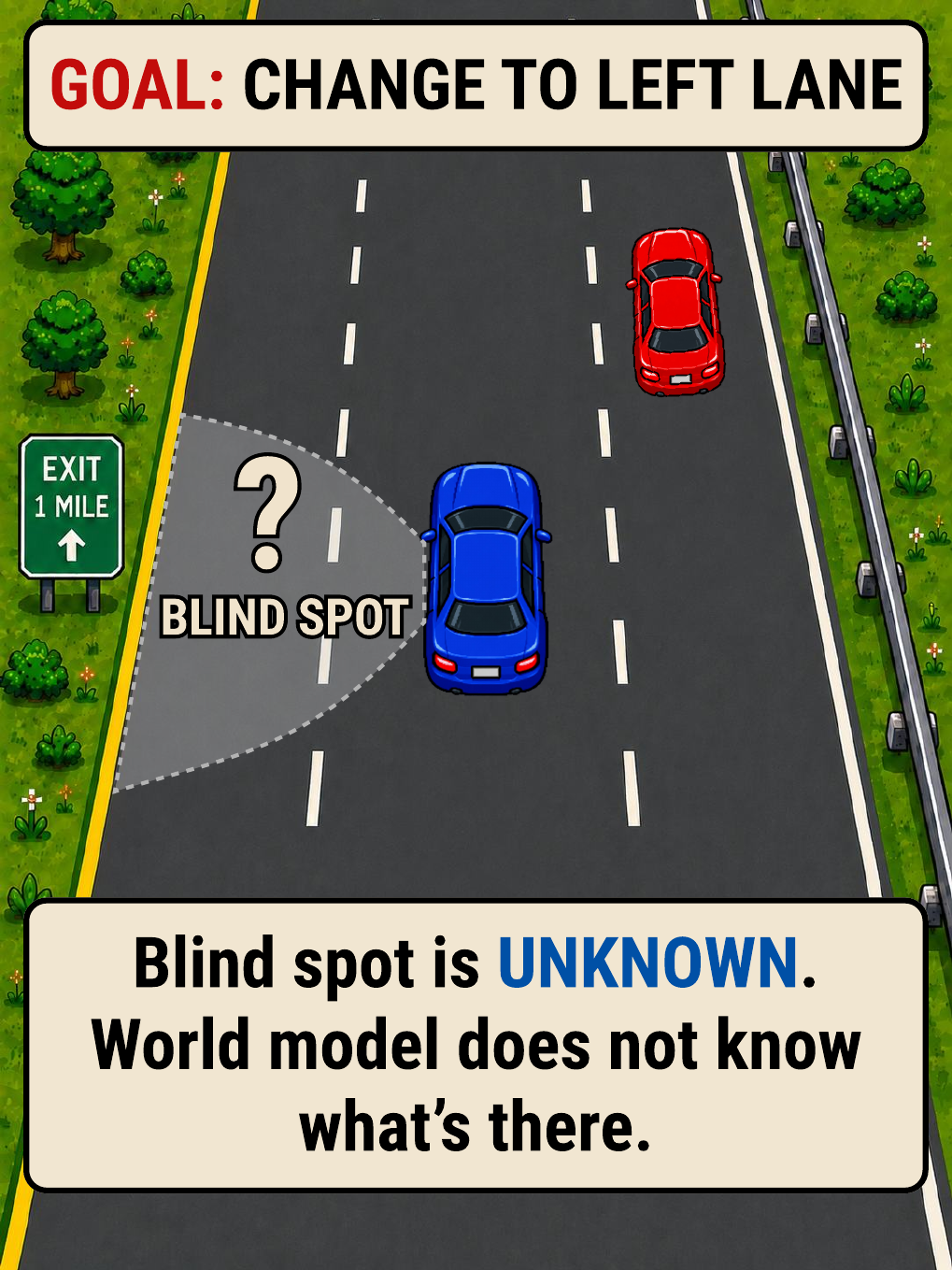}
        &
        \includegraphics[width=0.225\textwidth,height=4cm,keepaspectratio]{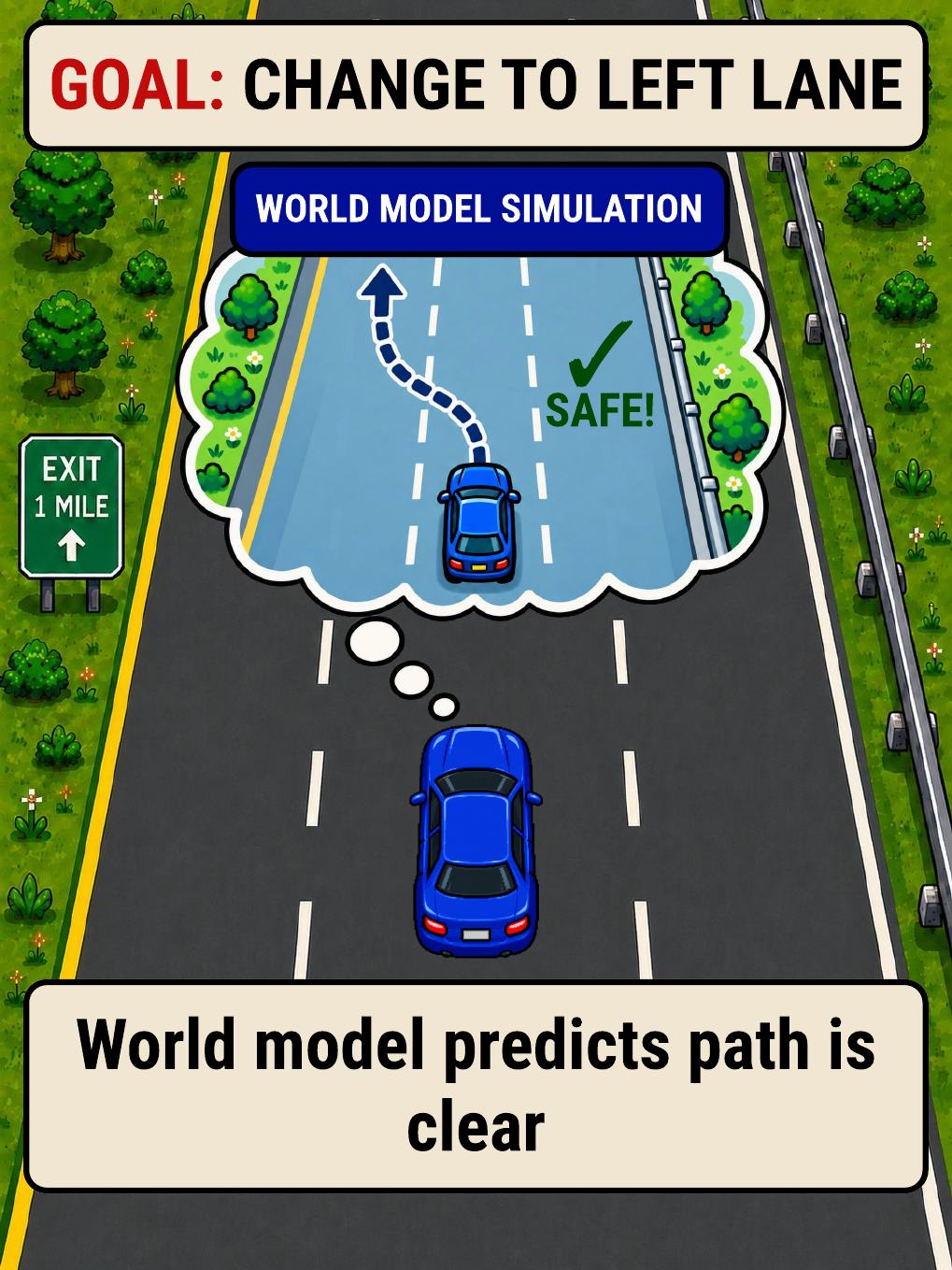}
        &
        \includegraphics[width=0.225\textwidth,height=4cm,keepaspectratio]{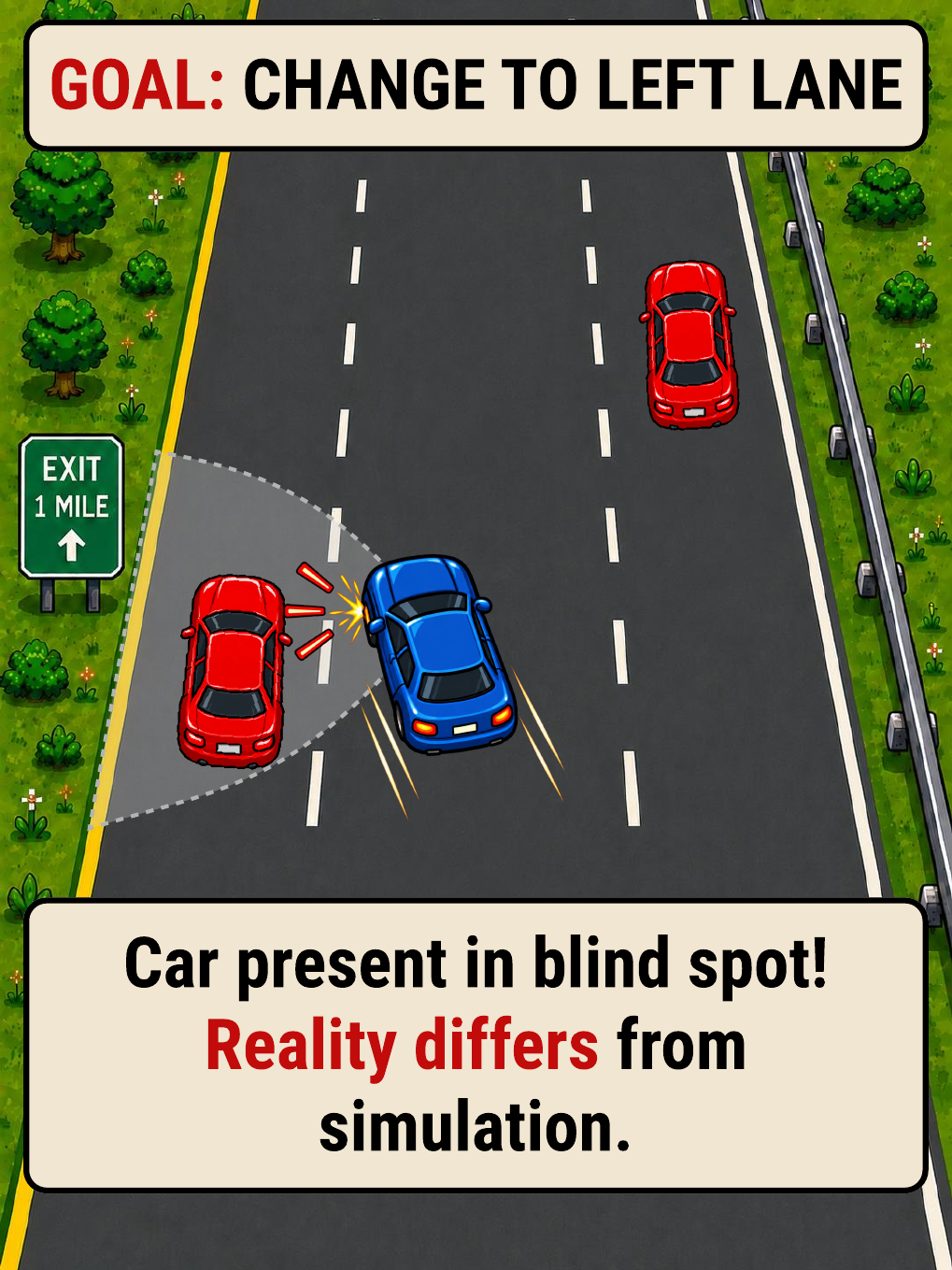}
        &
        \includegraphics[width=0.225\textwidth,height=4cm,keepaspectratio]{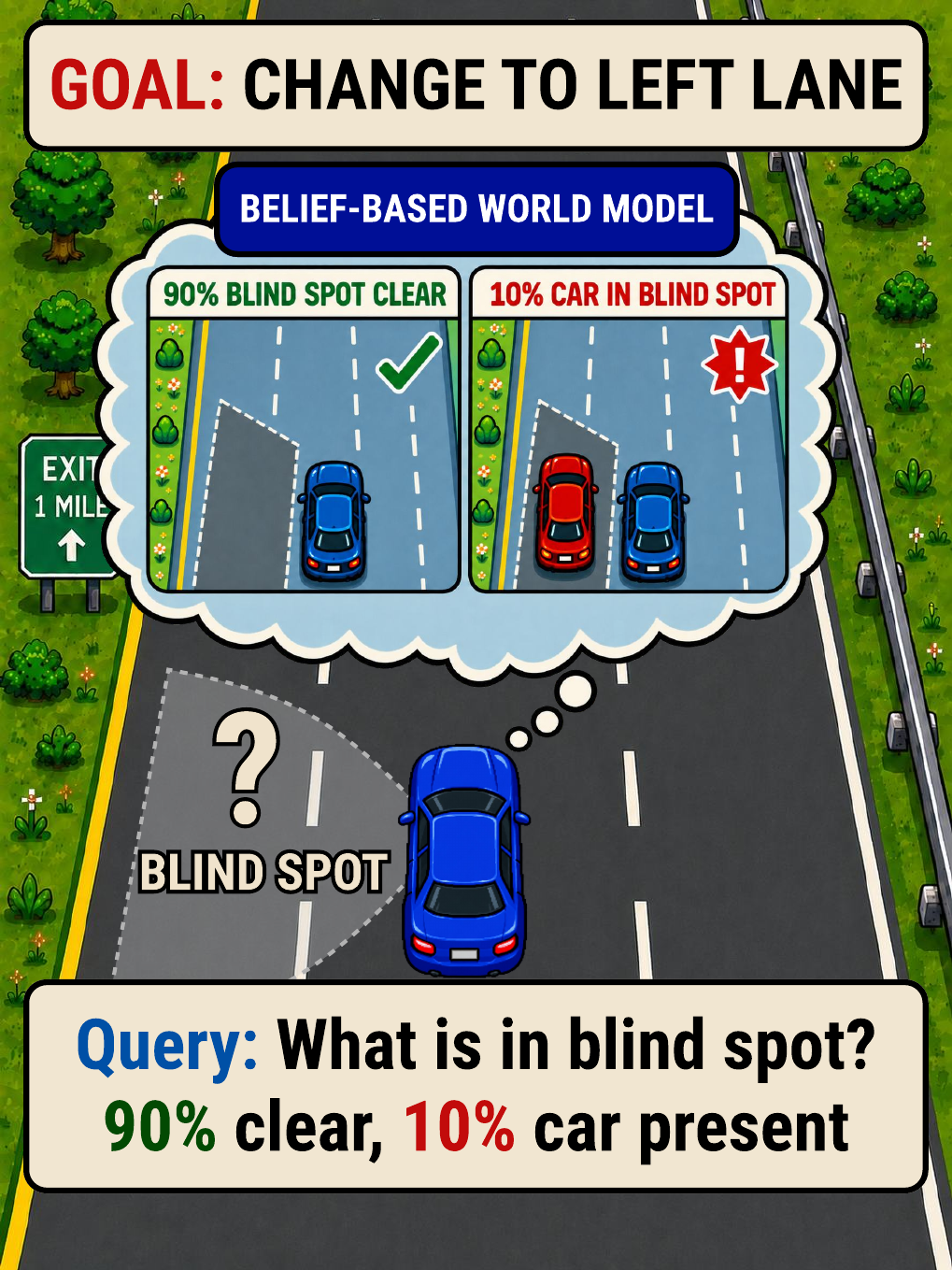}
    \end{tabular}

    \caption{\textbf{Simulation vs. belief for epistemic actions.} Consider a car attempting the \textit{pragmatic} action of changing to the left lane, where the blind spot is unknown. Before changing lanes, a simulation-based world model may sample futures where the lane change is safe (i.e., there is no car in the blind spot). The policy may then execute the action, only to encounter a car in its blind spot. With Belief-Based World Models, the policy can query for the current belief over the blind spot, and if there is uncertainty, it can take an appropriate \textit{epistemic} action (e.g., checking the blind spot) before choosing the appropriate pragmatic action.}
    \label{fig:motivating_example}
\end{figure}

Recent model-based control methods \citep{Maes2026LeWorldModelSE, zhou2025walle, hao-etal-2023-RAP, janner2022diffuser, schrittwieser2020muzero} allow the policy to interact with a world model through a simulation interface: given a representation of the current state and a candidate action (or action sequence), what future trajectory follows? We raise a broader question: \textbf{is simulation the only interface that a world model should expose to a policy?} While simulation may be sufficient for training model-free policies, we argue that it is inadequate for model-based control. Different types of actions require different information from a world model. Consider \textit{pragmatic} actions, which primarily advance the agent toward its goal (e.g., picking up a mug in order to drink coffee). Simulation is well-suited to these actions because it provides foresight by evaluating the consequences of acting. In contrast, \textit{epistemic} actions are information-gathering actions that primarily reduce uncertainty about the current state of the environment \citep{kaelbling1998planning, kirsh1994distinguishing}. In this setting, the policy benefits from a \textit{belief}---what is known and what is uncertain---about the current state to reason about whether epistemic actions are needed. Simulation can only expose uncertainty indirectly through variation across predicted futures, which may conflate uncertainty about the present state with stochasticity in the environment's state transition dynamics. Moreover, accurately recovering current-state uncertainty from simulation may require many samples and may still miss low-probability outcomes. Thus, we argue that the current-state belief should be exposed directly rather than reconstructed indirectly from future simulations (exemplified in Fig. \ref{fig:motivating_example}). 

Recently, large language model (LLM) agents have emerged as general-purpose policies in open-ended environments, from software engineering to robotics \citep{yang2024sweagent, workarena2024, pmlr-v162-huang22a, Ahn2022DoAI}. Given their general-purpose abilities, LLMs may be able to implicitly world model while planning actions. However, prior works have found that LLMs struggle with maintaining and updating beliefs in complex, long-horizon partially observable tasks \citep{Rahmani2026DebuggingCW, Jha2026EoG, zou2026T3_ICLR_Oral}. Recent world models for LLMs do not adequately alleviate this burden; they predominantly expose action-conditioned foresight while leaving belief estimation to the agent. The agent must therefore implicitly maintain uncertainty about the current state \textit{while simultaneously} reasoning about action selection, which is likely a suboptimal division of responsibilities. This motivates a modular setting: the trained LLM performs general reasoning and action selection, while a separate world model performs current and future state belief estimation. The remaining question is how these independent components should communicate. 

Separating belief estimation from action selection has precedence in decision-making under partial observability \citep{kaelbling1998planning} and has subsequently been explored in deep reinforcement learning (RL), where dedicated belief estimation can substantially improve decision-making \citep{hafner2020dreamer, pmlr-DVRL, pmlr-structured_world_belief, pmlr-v202-wang23p_believer, pmlr-forbes}. However, these approaches typically train the policy around a particular belief representation, and the control loop---the manner in which the policy and world model interact---is determined by the algorithm designer. This largely avoids the need to design an explicit interface between world model and policy. The LLMs considered in this work present a different setting. The LLM may be a pretrained, frozen, or even proprietary general-purpose reasoner that cannot readily be retrained around an arbitrary belief representation. At the same time, the agent's general reasoning abilities make a fixed control loop unnecessary: rather than specifying how the agent must use the world model, we can expose world model capabilities through an interpretable interface and allow the LLM itself to decide when they are useful. We therefore propose exposing belief in natural language, enabling an LLM to initiate queries about current or future states without being trained around a particular belief representation.

In short, we argue that the simulation interface of current world models for LLM agents should be augmented to explicitly expose beliefs over the current state. To achieve this, we introduce \textit{Belief-Based World Models} (BB-WMs), which should maintain a belief over the current state, update it as new observations arrive, and propagate it forward during action-conditioned simulation. This affords an LLM agent two complementary interfaces: natural language queries ask about the world model's current belief when reasoning about epistemic actions, while simulation evaluates the consequences of pragmatic actions. Before designing methods for learning scalable Belief-Based World Models, this paper focuses on a more fundamental question: \textbf{if a world model represents a belief over the state, does exposing this belief to the LLM agent improve decision-making?} To isolate this question, we intentionally design and study hand-crafted, benchmark-specific BB-WMs, abstracting away the separate challenge of learning accurate beliefs from experience.

\section{Related Works}
\label{sec:related-works}

\textbf{World Models for LLMs:} There have been efforts to equip LLM agents with separate world models that provide foresight into the consequences of candidate actions. WMA \citep{chae2025webagents_withWMs} finds that out-of-the-box LLMs cannot accurately simulate the outcome of actions in web interaction tasks, so they learn a separate world model to provide action-conditioned foresight. WALL-E \citep{zhou2025walle} replaces generative world models with a neurosymbolic world model, which checks the validity of an LLM-proposed action before it acts; if the action is predicted to be invalid, the LLM is asked to replan. DreamPhase \citep{hamidi2026dreamphase} learns a latent world model that generates hypothetical future observations from predicted future latent states, scores the generated futures with a learned value function, and distills feedback into natural language to condition a frozen LLM agent. These methods predominantly expose an action-conditioned simulation interface to the policy.

Interestingly, recent work by Qian et al. \citep{qian-etal-2026-current} shows that providing an LLM agent with a ground-truth simulator alone does not translate into improved decision-making at test-time; agents often fail to invoke simulation when useful, misuse its outputs, or even degrade when simulation is enforced. Their analysis suggests that a key challenge is in when an agent decides to query a world model and how it integrates the resulting information into its reasoning. These findings motivate our efforts to extend world models with belief modeling capabilities, allowing the LLM to benefit from a different type of information.

\textbf{Belief Representation in LLM Agents:} Some works have sought to improve LLM agents by maintaining estimates of the current environment state rather than relying solely on an agent's ability to reason over the interaction history. Approaches such as Statler, QuBE, StateAct, and ABBEL \citep{yoneda2024statler, kim-etal-2024-qube, rozanov2024stateactstatetrackingreasoning, Lidayan2025ABBELLN} maintain compact textual or structured representations of task-relevant state that are updated as new observations become available. These representations emphasize a single estimate or summary of the current state rather than explicitly preserving uncertainty over many possible states. Some recent work has begun to model this uncertainty directly. BeliefMem \citep{Liao2026BeliefMem} retains multiple candidate conclusions together with probabilities, allowing competing interpretations of past evidence to coexist and updating them as additional evidence is observed. Agent-BRACE \citep{singh2026agentbracedecouplingbeliefsactions} more directly adopts the POMDP notion of belief for LLM agents, representing the current state uncertainty and conditioning the policy on this structured belief. These works demonstrate growing interest in explicitly representing uncertainty about the current state and making this information available to the agent during decision making. \textbf{However, these approaches are not world models.} They focus on estimating and maintaining the environment's \emph{current} state but do not provide an action-conditioned simulation interface for reasoning about the future. BB-WMs seek to bridge simulation with belief estimation: the policy can directly access the world model's belief when reasoning about uncertainty in the current state, while continuing to obtain foresight through simulation.





\begin{figure}[!tbp]
    \centering

    \begin{subfigure}[t]{0.30\textwidth}
        \centering
        \includegraphics[width=\linewidth,height=3.2cm,keepaspectratio]{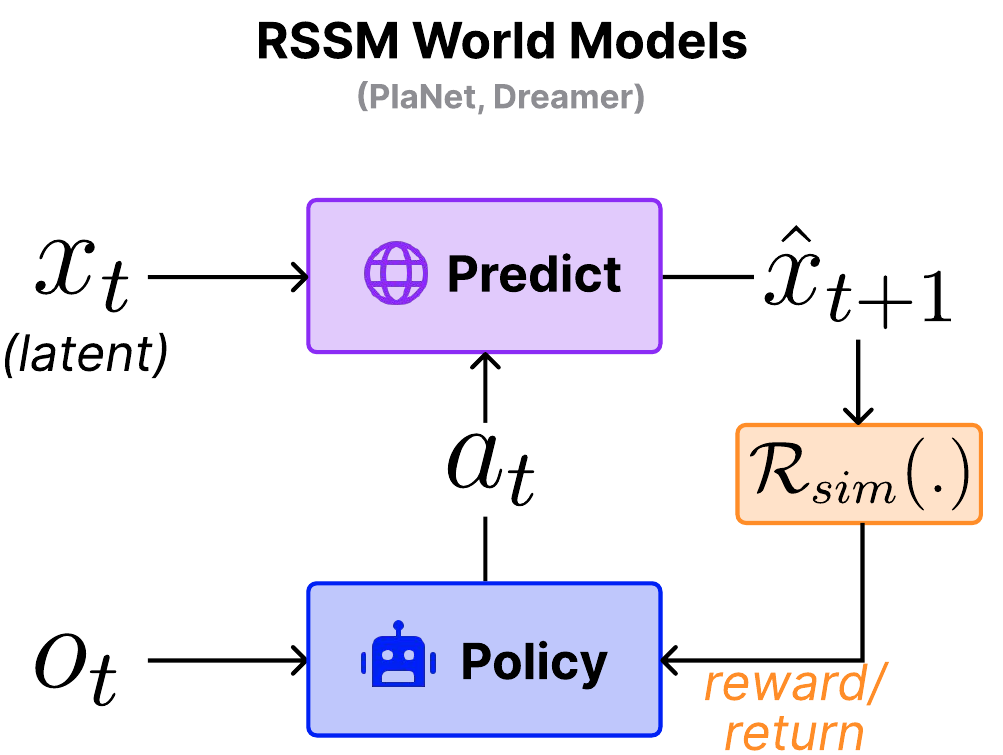}                
    \end{subfigure}
    \hfill
    \begin{subfigure}[t]{0.30\textwidth}
        \centering
        \includegraphics[width=\linewidth,height=3.2cm,keepaspectratio]{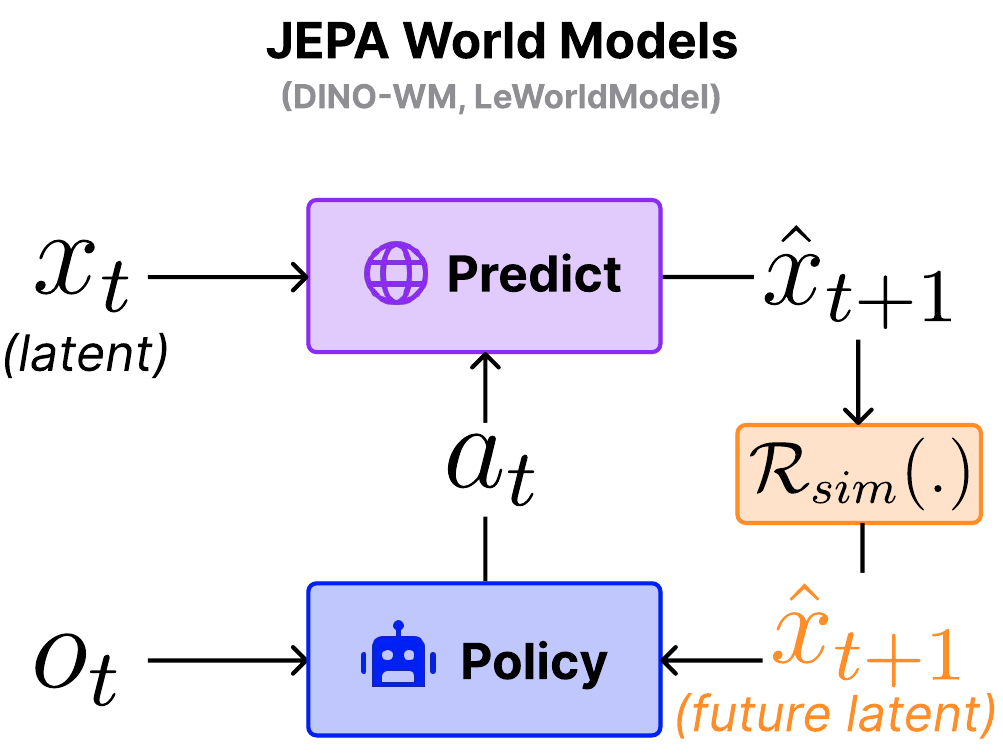}                        
    \end{subfigure}
    \hfill
    \begin{subfigure}[t]{0.30\textwidth}
        \centering
        \includegraphics[width=\linewidth,height=3.2cm,keepaspectratio]{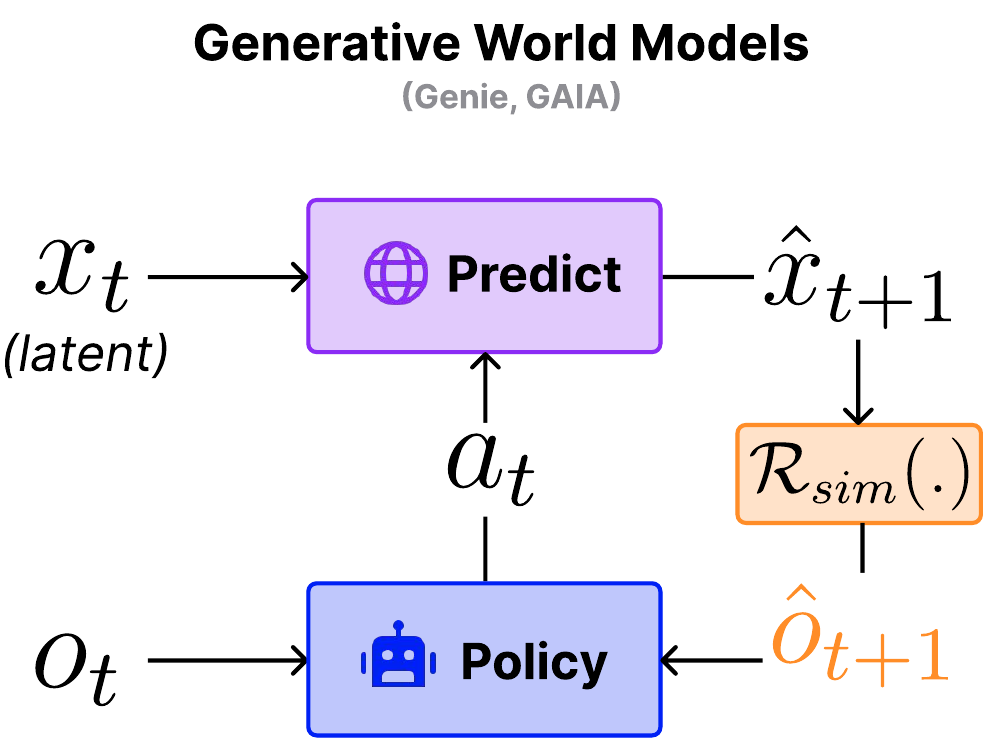}                        
    \end{subfigure}

    \vspace{0.25cm}

    \begin{subfigure}[t]{0.30\textwidth}
        \centering
        \includegraphics[width=\linewidth,height=3.2cm,keepaspectratio]{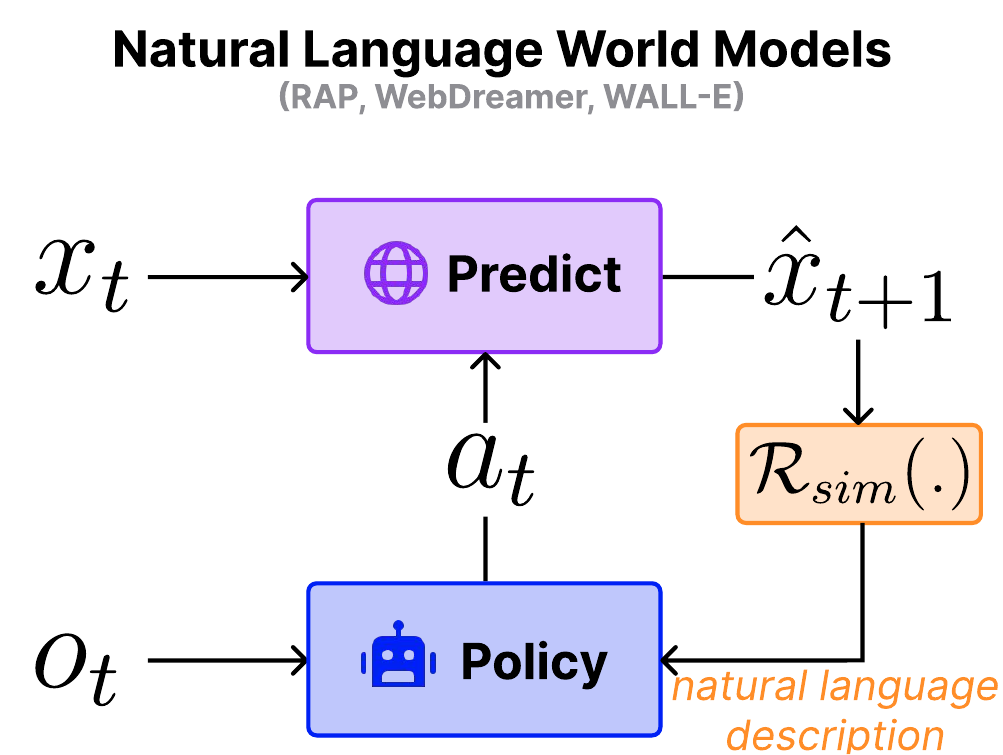}                    
    \end{subfigure}
    \hfill
    \begin{subfigure}[t]{0.66\textwidth}
        \centering
        \includegraphics[width=\linewidth,height=3.2cm,keepaspectratio]{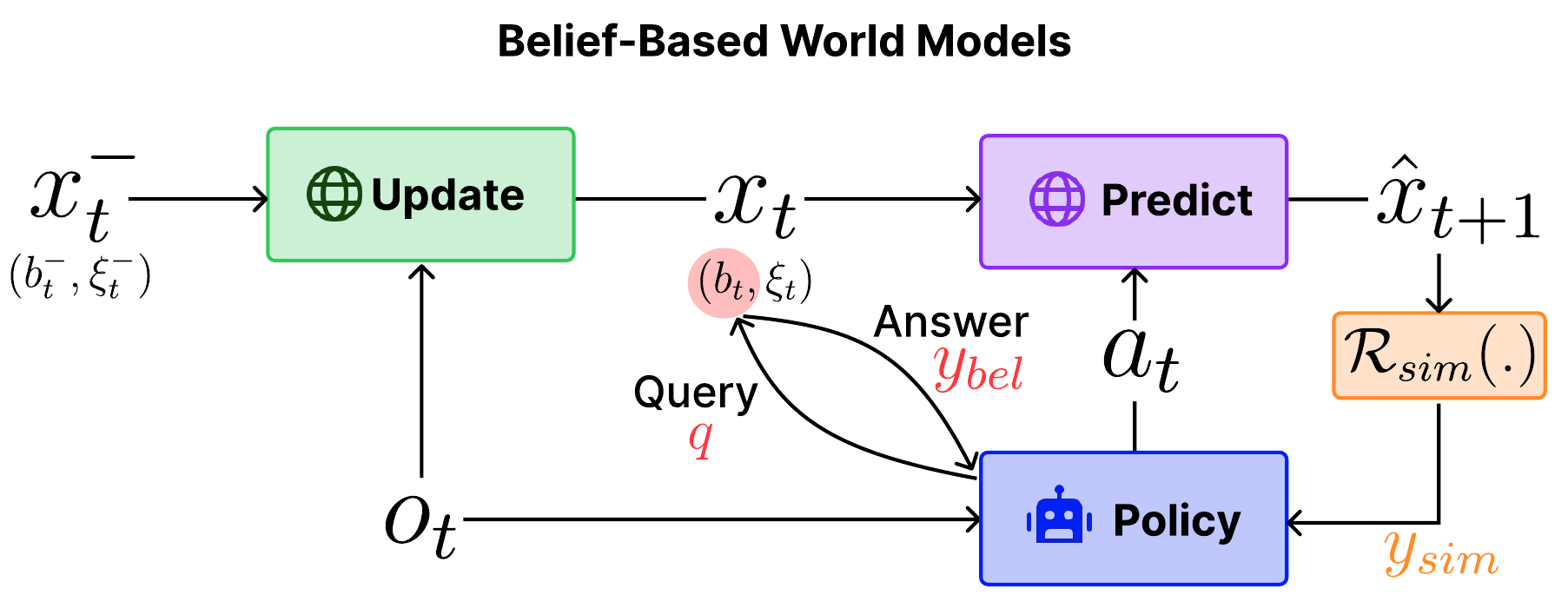}                        
    \end{subfigure}

    \caption{\textbf{BB-WMs vs. prior world models}. Prior world models for model-based control use simulation to give the policy foresight into the future. Recurrent state-space model (RSSM) world models \citep{hafner2020dreamer, hafner2018planet} represent the state as a latent and respond to the policy with the predicted (cumulative) reward of an action sequence. JEPA world models \citep{dinowm, Maes2026LeWorldModelSE} respond to the policy directly with the predicted latent state. Generative world models \citep{GenieGoogle, Hu2023GAIA1AG} instead decode the predicted latent into a predicted observation. Natural-language world models \citep{hao-etal-2023-RAP, zhou2025walle, Gu2025WebDreamer}, regardless of their internal state representation, respond to the policy in natural language (either free-form or symbolic). \textbf{In contrast, BB-WMs also allow the policy to ask about the current belief.}}
    \label{fig:wm_comparisons}
\end{figure}

\section{Belief-Based World Models}
\label{sec:bb-wm}

\textbf{Preliminaries:} An agent, at any timestep $t$, receives an \textit{observation} $o_t$, takes \textit{action} $a_t$ in the environment, and receives the subsequent observation $o_{t+1}$. The agent selects actions in order to achieve some goal $g$. A world model maintains some \textit{internal representation} $x_t$ of the environment's true latent state $s_t$ based on the agent's interaction history. Most world models used for model-based control expose a similar interface to the policy. Specifically, the policy queries the world model with $K$ candidate actions $\{a_t^{(i)}\}_{i=1}^K$ (in practice, the policy may query with a sequence of actions). The world model predicts (or samples) the representation of the future state after taking the action: $\hat{x}^{(i)}_{t+1} = Predict(x_t, a_t^{(i)})$. The predicted internal state need not itself be exposed to the policy. We denote by $R(\hat{x}^{(i)}_{t+1})$ the readout of the predicted future, which may take many different forms, including a latent representation, natural-language description, symbolic state, or generated observation. The policy uses these readouts to determine which action to execute in the real environment. Prior methods may differ in terms of their internal representation $x_t$ and the response given $\mathcal{R}(\hat{x}_{t+1})$ to the policy, but this simulation interface underlies many world model families, which we show in Fig. \ref{fig:wm_comparisons}.

\textbf{Belief-Based World Models:} In a BB-WM (see bottom right of Fig. \ref{fig:wm_comparisons}), the internal representation $x_t=(b_t, \xi_t)$ includes an explicit belief $b_t$ over the environment's current latent state $s_t$, with $\xi_t$ denoting any additional information. The BB-WM specifies an interface that allows the belief to be directly exposed to the policy. Specifically, given all past observations and actions $h_t = (o_{1:t}, a_{1:t-1})$, the belief is $b_t \equiv P(s_t | h_t)$. Before executing $a_t$, the \textit{prior} representation of the next timestep is $x^-_{t+1} = (b^-_{t+1}, \xi^-_{t+1}) = Predict (x_t, a_t)$. After executing $a_t$ and observing the resulting $o_{t+1}$, the \textit{posterior} representation is $x_{t+1} = (b_{t+1}, \xi_{t+1}) = Update(x^-_{t+1}, o_{t+1})$. Then, the policy has access to two interfaces. 

\textbf{Simulation interface:} For a candidate action $a_t^{(i)}$, the world model predicts the resulting next state $\hat{x}^{(i)}_{t+1}$. The policy receives a policy-facing readout: $y_{sim}^{(i)} = R_{sim}(\hat{x}^{(i)}_{t+1})$.

\textbf{Belief-query interface:} Unlike current world modeling methods, with a BB-WM the policy can directly query the current posterior belief \textbf{without} supplying a hypothetical action. Specifically, the policy can ask some query $q$ over belief $b_t$ through the interface: $y_{bel} = R_{bel}(b_t,q)$.

Before taking an action, the policy may issue multiple queries across both interfaces, and the responses can be used as conditioning for the policy to decide the next action.

\section{Method}

Recall our research question: ``if a world model represents a belief over the state, does exposing this belief to the LLM agent improve decision-making?" To isolate the effect of exposing the belief to the agent, we deliberately abstract away the separate challenge of learning an accurate BB-WM by studying simple text-based environments and leveraging prior knowledge about the game environment. This allows us to hand-specify the task-relevant state space, the prior belief prediction $Predict(.)$, and the posterior belief update $Update(.)$, instead of needing to learn them from environment interactions. Because the hand-designed state space is semantic and interpretable, we can define a natural-language belief readout $R_{bel}(b_t,q)$, allowing a pretrained LLM policy to query the BB-WM without finetuning. \textbf{Our goal is not to propose hand-crafted world models as a scalable solution, but to establish whether explicit belief access is useful in the first place.} We describe our benchmark-specific instantiations next.

\subsection{ALFWorld}

ALFWorld \citep{ALFWorld20} is a text-based game where an agent completes household instructions in a simulated home. The agent must locate one or more target \textit{objects} among a set of \textit{receptacles} (cabinets, drawers, fridge, etc.), manipulate the objects (take, clean, heat, cool, etc), and place them at a specified receptacle. The environment is partially observed: objects are randomly initialized to specific receptacles, and receptacle contents are hidden until the agent navigates to them.

\textbf{Prompt:} An LLM agent is used as the policy. In the initial prompt, we provide environment-specific details, the agentic framework instructions, available actions, the WM query interface, and an in-context learning (ICL) example. The ICL example show the agent an end-to-end example of completing a task from the task-type and involves a WM belief query. Then, the goal task and initial observation are provided. Details and examples are in Appendix \ref{app:alfworld_agent_prompt}.

\textbf{State Space:} The state space contains \textit{deterministic} components and \textit{probabilistic} components. Deterministic components include things like agent location and inventory. The only probabilistic component is a belief over object locations. It is represented as a categorical distribution over all receptacles in the environment. More details on the state space construction are in Appendix \ref{app:alfworld_state}.

\textbf{Belief Update:} Deterministic components are updated using rule-based logic, parsed from ALFWorld environment observations. The probabilistic component is seeded from a uniform prior over possible receptacles for each object, which is prior knowledge specified by ALFWorld's game engine (see Appendix \ref{app:alfworld_probabilistic_state} for more). Our belief update follows simple presence/absence renormalization: finding an object in a receptacle collapses the belief to a point mass; if not found, we zero that receptacle's belief and renormalize the belief over the remaining unsearched receptacles.

\textbf{Simulation interface:} We use WALL-E \citep{zhou2025walle}, which implements a rule-based mechanism to check if the contemplated action is valid (given the current static state) and responds to the agent in natural language (either valid or invalid with feedback). If WALL-E returns the action is valid, the agent executes it. Else, the agent re-plans the action conditioned on the feedback.

Note that WALL-E is an incomplete next-state predictor. It does not actually predict the complete next state, and it does not represent a belief. Regardless, we intentionally adopt it because it is simple and allows us to test the benefits of the belief-query interface in the BB-WM.

\textbf{Belief-query interface:} The world model's belief is exposed to the agent through a \textit{query} action. To access the probabilistic state, the agent can ask \textit{where is <object>}, and the WM responds with a list of receptacles with non-zero probabilities (ranked by their belief). The agent can also access the static state through other queries. See Appendix \ref{app:alfworld_belief_query_interface} for more details on the interface.


\subsection{ScienceWorld}
\label{sec:method-scienceworld}

ScienceWorld \citep{wang-etal-2022-scienceworld} is a text-based game in which an agent performs elementary-science experiments in a fixed ten-room house. We evaluate on 24 test task types (e.g., changing states of matter, growing a plant, mixing chemicals). Relative to ALFWorld, ScienceWorld requires more common-sense reasoning from the agent. Similar to ALFWorld, uncertainty is limited to the location of randomly initialized task-relevant objects. However, we observe that the possible locations of task-relevant objects is far less varied compared to ALFWorld.

We follow ALFWorld's prompt structure; details and examples are in Appendix \ref{app:scienceworld_agent_prompt}. The deterministic components of the world model's state space are detailed in Appendix \ref{app:scienceworld_deterministic_state}. The probabilistic component operates similarly to what was done for ALFWorld (see Appendix \ref{app:scienceworld_probabilistic_state}). Rule-based logic is used to update deterministic components of the state space. Similar to ALFWorld, belief over object locations is updated using presence/absence renormalization.

\textbf{Simulation Interface:} Since there is no publicly available implementation of WALL-E for ScienceWorld, we opt for an oracle version of WALL-E. We use the environment to check if the action is valid or invalid. If valid, the action is executed, and if invalid, the agent is given generic feedback and asked to retry (up to the same retry budget used in WALL-E). We refer to this as an oracle because we are guaranteed to always get the valid/invalid action prediction correct.

\textbf{Belief-query interface:} To access the probabilistic state, the agent can ask \textit{where is <object>}, and the WM responds with a list of rooms ranked by their belief, and the container of the object if known. The agent can also query the deterministic state. See Appendix \ref{app:scienceworld_belief_query_interface} for more specifics.

\subsection{BabyAI}

BabyAI \citep{chevalier-boisvert2018babyai} is a text-based game in a grid world environment, where the agent uses simple navigation commands (e.g., go forward, turn left/right, pick/drop) to complete a pre-specified goal. We evaluate on 4 task types from the BALROG benchmark \citep{paglieri2025balrog} (details in Appendix \ref{app:babyai}). Relative to our other benchmarks, BabyAI requires spatial reasoning and lower-level planning, since the agent cannot issue semantic commands like ``go to kitchen''. BabyAI adds partial observability through a field-of-view (FoV) mechanism, where only cells in the agent's FoV are observed. We generally follow ALFWorld's prompt structure, except there are no ICL examples provided (see Appendix \ref{app:babyai_agent_prompt} for more). The deterministic and probabilistic components are detailed in Appendix \ref{app:babyai_state}. Since there is no WALL-E implementation for BabyAI, we create our own (detailed in Appendix \ref{app:babyai_walle}). To access the probabilistic state, the agent can ask for an object's location or a map of the grid. The agent can also query deterministic components. See Appendix \ref{app:babyai_belief_query_interface} for more on the query interface.

\section{Experiments}

Our goal is to study whether exposing a belief over the current state to an LLM agent improves its decision-making. We therefore perform a controlled study in which the LLM agent is frozen and variants differ only in the information exposed by the world model. Our objective is not to maximize benchmark performance, but to isolate the relative benefit of belief access. For this reason, we also do not compare to concurrent methods for belief tracking (referenced in Sec. \ref{sec:related-works}).

\textbf{Methods Evaluated:} On each benchmark, we evaluate three LLMs (Llama-3.1-8B-Instruct \citep{grattafiori2024llama3herdmodels}, Qwen3-14B \citep{yang2025qwen3technicalreport}, and Sonnet 4.6) on two agentic frameworks (ReAct \citep{yao2023react} and ReflAct \citep{kim-etal-2025-reflact}). For each agent, we evaluate several different world models. In \textit{Belief}, we allow the agent to only ask queries about the current belief state (no simulation). In \textit{WALL-E}, we pair the agent with WALL-E's action-conditioned next-state predictor (no belief state, just a valid/invalid action prediction). In \textit{BB-WM}, we compose the belief modeling with WALL-E, allowing the agent to both do simulation and ask questions on the current belief state. 

\textbf{Evaluation Budget:} In ALFWorld/BabyAI, the agent can take up to 30/64 environment actions to achieve the task, following. ScienceWorld varies the step budget per task family. In both benchmarks, we terminate execution early if the agent seems to be derailed; specifically if it issues 10 consecutive invalid actions or 10 consecutive world model queries.

\textbf{Metrics:} An effective agent should complete the most number of tasks in the fewest number of steps. We use \textit{reward-within-budget} to measure both. The average reward obtain so far is reported as a function of the fraction of steps taken out of the total action budget. ALFWorld's and BabyAI's return is binary (0/1) depending on task success and ScienceWorld's reward is a score from [0,1], allowing partial rewards in intermediate steps for reaching certain milestones.

\begin{figure}[tbp]
    \centering

    \includegraphics[width=\linewidth]{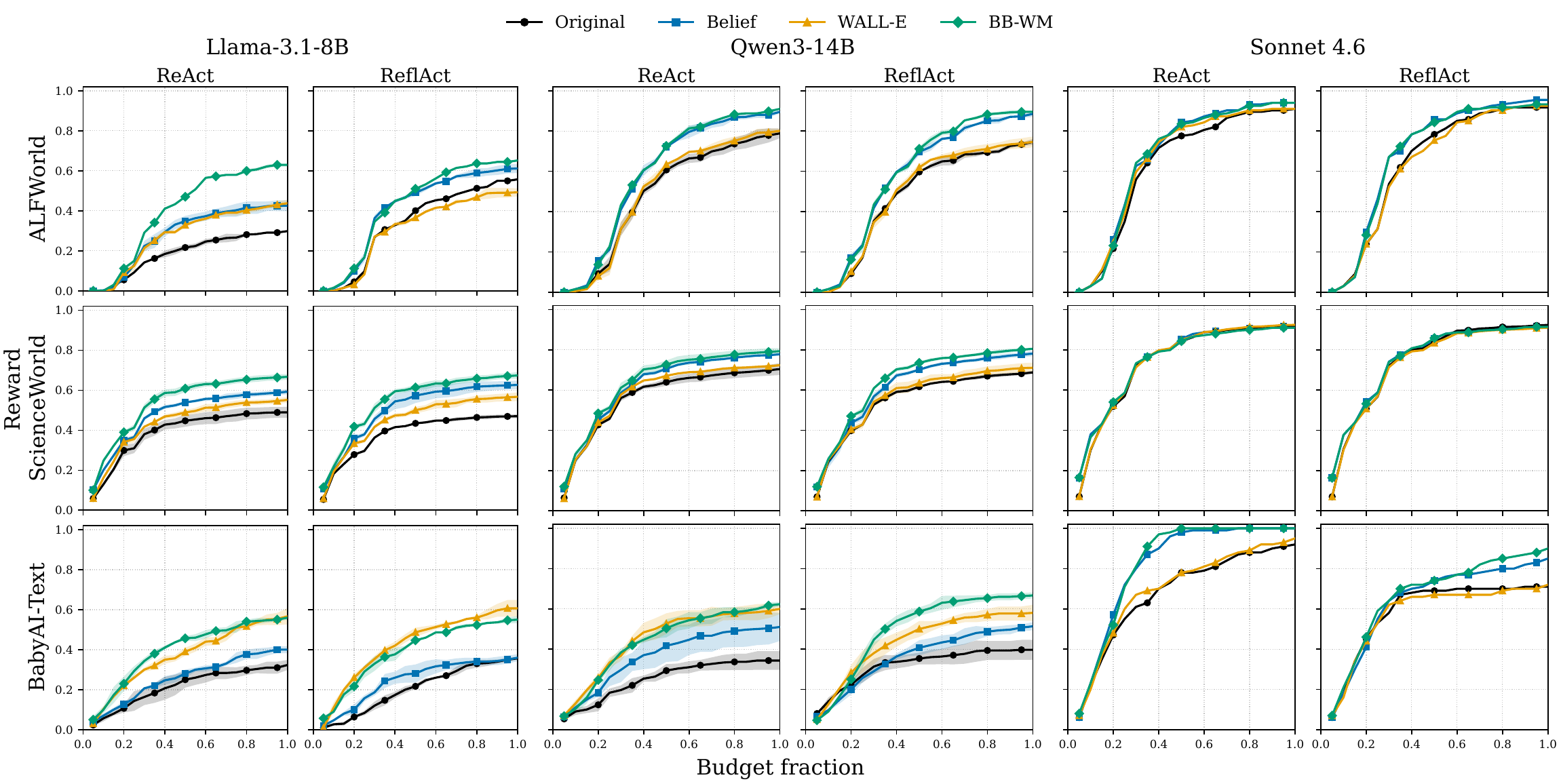}

    \caption{\textbf{Benchmark results.} Each LLM is run on two agentic frameworks (ReAct and ReflAct). We add a \textit{Belief} variant, where the agent can query about the current belief state (no simulation). In \textit{WALL-E}, the agent is paired with an action-conditioned next-state predictor (no belief state). In \textit{BB-WM}, we compose WALL-E and belief-states together, allowing for both simulation and queries on the current belief state. Our metric is reward-within-budget, which measures reward obtained as a function of environment steps. Higher is better. Mean and standard deviation over three runs (except for Sonnet) are reported. \textbf{Takeaway:} BB-WM improves over the base agent in terms of efficiency and performance. BB-WM is generally better than beliefs or WALL-E alone.}

    \label{fig:alfworld_main_results}
\end{figure}

\subsection{Results}
\label{sec:main-results}

We report results on the benchmarks in Fig. \ref{fig:alfworld_main_results}. We make the following observations:

(1) \textbf{Adding \textit{BB-WM} helps the base agent in all cases}, both in performance and efficiency. In general, \textit{BB-WM} is superior to the other variants (\textit{WALL-E} and \textit{Belief}).

(2) Enabling belief queries (represented by \textit{Belief}) and simulation queries (represented by \textit{WALL-E}) have a \textbf{complementary effect: they address different shortcomings in the base agent}. For example, on ALFWorld the Llama ReAct agent enjoys performance gains under each method individually, which compound when they are combined in the \textit{BB-WM}. In other cases, \textit{WALL-E} does not improve the base agent's performance (observe, Qwen3 ReflAct on ScienceWorld), but when integrated into the \textit{BB-WM}, the agent outperforms the \textit{Belief} case. 

(3) \textbf{The LLM benefits from beliefs if the task is hard relative to the LLM's capabilities.} Sonnet, by itself mostly saturates ALFWorld and ScienceWorld. Nevertheless, we notice consistent, small efficiency gains over the base agent when incorporating beliefs. On BabyAI, however, Sonnet has not saturated the benchmark, and adding WALL-E does not make much of an impact, likely because Sonnet does not tend to perform invalid actions. Adding beliefs (either with \textit{Belief} or \textit{BB-WM}) bring substantial improvements in terms of efficiency and performance. For easier tasks, an LLM of sufficient scale may be able to implicitly track state (even with uncertainty) and rarely performs invalid actions, which constrains the benefits of any world modeling efforts; for more difficult tasks, an LLM benefits from access to beliefs, and the benefits tend to compound when adding foresight through simulation-based mechanisms.

\subsection{Memory vs. Belief}

\begin{figure}[tbp]
    \centering

    \includegraphics[width=\linewidth]{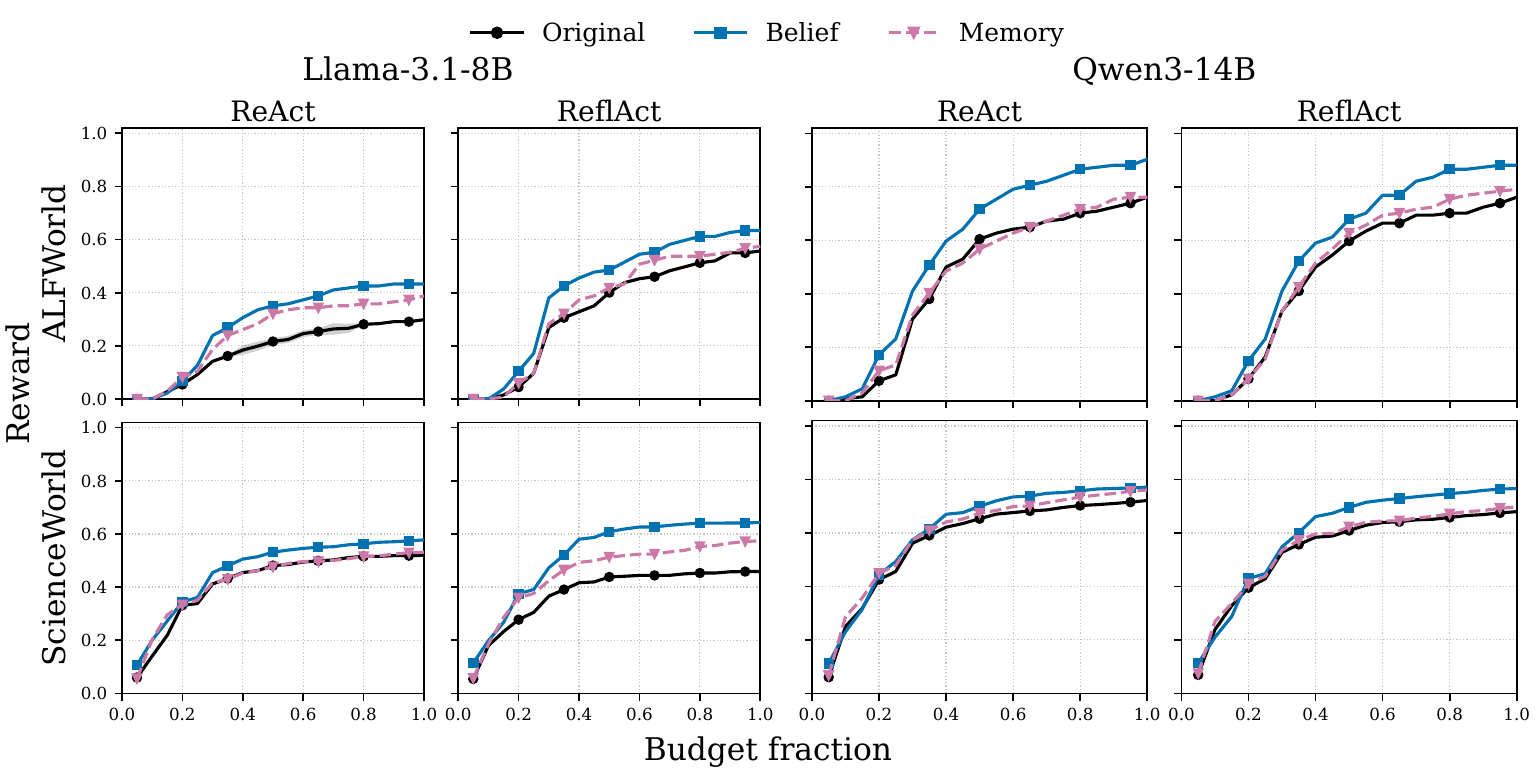}

        \caption{\textbf{Memory vs. belief.} We compare \textit{Belief} against a \textit{Memory}-only variant that tracks information from past observations but does not maintain uncertainty over unobserved object locations. \textbf{Takeaway:} Memory alone does not explain the gains from belief access; explicitly representing a belief over uncertain components of the state improve agent decision-making.
    }

    \label{fig:memory_belief_results}
\end{figure}

Our world model maintains deterministic information extracted from past observations, together with a belief over uncertain components of the current state. The deterministic component can be viewed as within-episode \textit{memory}: it records information acquired earlier in the trajectory. Since prior work has shown that explicit memory can improve LLM agents \citep{yoneda2024statler, hu-etal-2025-hiagent}, a natural alternative explanation is that the gains from \textit{Belief} arise primarily from externalizing this memory rather than from representing uncertainty.

To evaluate these effects, we introduce a \textit{Memory}-only ablation on ALFWorld and ScienceWorld for the Llama and Qwen agents. \textit{Memory} retains the same deterministic state as \textit{Belief}, but removes the belief over unobserved object locations. Accordingly, queries about the observed state are unchanged, while a \textit{where is <object>} query for an unobserved object only reports that the object has not yet been observed. Thus, \textit{Memory} provides access to what the agent has already seen from the trajectory, whereas \textit{Belief} additionally represents what remains possible about the unobserved state.

Results are shown in Fig. \ref{fig:memory_belief_results}. Memory provides gains over the base agent in some settings, re-affirming that state tracking is useful. However, \textit{Belief} consistently improves over \textit{Memory}. These results indicate that the \textbf{gains from belief access cannot be explained by deterministic memory alone:} a belief over uncertain components of the state provides additional decision-relevant information.

\subsection{Non-Uniform Priors}

In our main experiments, the belief over an object's initial location is uniform over its valid receptacles, reflecting the initialization logic of the underlying benchmarks. In this setting, knowing the belief support---which receptacles remain possible---captures most of the useful information, since these locations are equally likely. In more realistic partially observed environments, however, some states may be substantially more likely than others. We therefore ask whether an LLM agent benefits from access to the \emph{probabilities} of a belief, beyond simply knowing its support.

\begin{figure}[tbp]
    \centering

    \includegraphics[width=\linewidth]{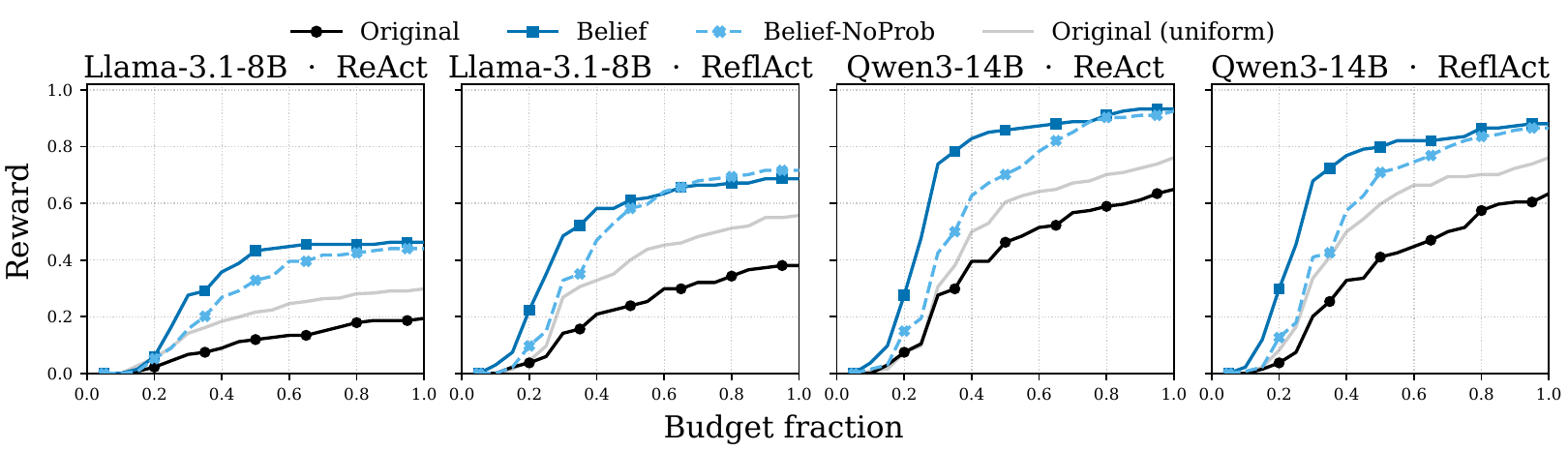}

    \caption{\textbf{Non-uniform priors:} We modify ALFWorld so that target objects are initialized across valid receptacles according to a non-uniform distribution rather than uniformly. We compare the base agent, \textit{Belief}, and \textit{Belief-NoProb}, which exposes the same possible locations as \textit{Belief} but omits the corresponding probability values. \textbf{Takeaway:} Explicit probability information improves efficiency: the agent benefits from not just knowing which locations the object may exist at (\textit{Belief-NoProb}), but \textit{how likely} a location is (\textit{Belief}).}

    \label{fig:non_uniform_prior_results}
\end{figure}

To study this question, we modify ALFWorld's object-initialization process. For each object type, we define a Zipf (or zeta) distribution ($\alpha=1.25$) over its valid receptacles, which we use to sample the task-relevant object's initial location. We use a Zipf distribution because it introduces a simple heavy-tailed ranking over otherwise-valid locations: a small number of receptacles are substantially more likely, while all valid receptacles retain non-zero probability. The BB-WM is given the corresponding prior and updates it as observations eliminate possible locations. We additionally introduce \textit{Belief-NoProb} to isolate the value of explicit probability information. This variant is identical to \textit{Belief}, except that the world model omits numerical probabilities when answering location queries and randomizes the location order in the response to remove ordinal information. Thus, the agent still knows which receptacles are possible, but does not receive the probability mass assigned to each one. Results are shown in Fig.~\ref{fig:non_uniform_prior_results}.

(1) \textbf{Non-uniform initialization makes the base agent less effective.}
Although the tasks, environments, and set of valid object locations are unchanged, altering the initialization distribution reduces the performance of the base agent. Providing access to the belief largely recovers the performance observed under the original initialization process (in Sec. \ref{sec:main-results}).

(2) \textbf{Probability information improves search efficiency.} \textit{Belief} and \textit{Belief-NoProb} reach similar performance by the end of the action budget, but \textit{Belief} achieves substantially higher reward earlier in the trajectory. Thus, the benefit of belief access is not limited to identifying which states remain possible: the agent uses the probability assigned to those states to prioritize more likely locations and solve tasks using fewer actions.

\subsection{Qualitative Analysis}

BB-WMs help the agent generalize to challenging test-time situations. We see this with an ALFWorld task that the Sonnet ReAct agent failed with WALL-E but completed with BB-WM. The task is to \textit{heat some egg and put in} \texttt{garbagecan}. The main difficulty in this task is that the egg gets randomly initialized to the \texttt{garbagecan}, which likely represents a novel setting for Sonnet (in the real world, eggs are not commonly found in the \texttt{garbagecan}, especially those that need to be heated). In the WALL-E trajectory (see Appendix \ref{app:alfworld_trajectory}), the agent exhausts the action budget by checking many receptacles (fridge, countertops, \textcolor{violet}{shelves}, \textcolor{violet}{cabinets}, \textcolor{violet}{drawers}, \textcolor{violet}{stoveburners}). Critically, the ALFWorld game engine will \textit{never} initialize an egg into the \textcolor{violet}{violet}-colored receptacles, but WALL-E cannot expose this world knowledge to the LLM agent. When coupled with the BB-WM (see Appendix \ref{app:alfworld_trajectory}), the agent starts by querying for possible egg locations, and the BB-WM, due to its prior, only responds with valid initial locations. The agent checks the more sensible receptacles first (fridge, countertop, microwave), then re-queries the BB-WM for an updated receptacle list, and finally checks the \texttt{garbagecan} to find the egg, heat it, and complete the task. 


\section{Conclusion}

This work shows that belief access improves LLM agent decision-making and often compounds with simulation-based world modeling, motivating \textit{Belief-Based World Models} (BB-WMs), where current-state belief and future-state prediction provide distinct and complementary information to an agent. This work has two important limitations. First, our BB-WMs are hand-crafted and benchmark-specific. Thus, our results establish that \emph{accurate beliefs are useful when exposed to an LLM policy} but do not address how such beliefs should be learned in realistic, open-ended settings. Second, our simulation interface is based on WALL-E-style action-validity prediction. This is a relatively limited form of simulation and does not propagate the current belief forward into a belief over hypothetical future states. Both limitations are important directions for future work, and our results indicate that addressing them could yield meaningful benefits in agent decision-making under partial observability.

\subsection*{AI use statement}

In this work, we used generative AI tools for creating or modifying scientific figures or images, creating or editing software code, summarizing or analyzing existing literature, sourcing/searching for information, editing a research paper to improve readability, and identifying relevant literature.

We have not used generative AI tools to help develop theoretical models or conceptual frameworks, propose or refine hypotheses, design or provide feedback on research methodology or experiments, implement methods, support qualitative and thematic data analysis, interpret results. The following are not applicable to this work: generate synthetic data sets, formulate mathematical claims, provide critical ingredients for proving mathematical claims, assist in the writing of proofs, assist with translation, clean and reformat dataset.

We have reviewed all AI-assisted work. Figures and images were checked for correctness and readability. AI-generated code was tested for correctness. Sources raised in literature reviews were manually reviewed and read to obtain a more complete understanding. We take responsibility for the final content of this work, including text, claims or artifacts produced with the aid of generative AI.

\bibliography{iclr2027_conference}
\bibliographystyle{iclr2027_conference}

\appendix

\section{BB-WM Instantiation for ALFWorld}

\subsection{General Agent Prompt}
\label{app:alfworld_agent_prompt}

We provide the agent prompt used for ALFWorld below. We separate the prompt
into benchmark instructions, agent instructions, the available action space,
the in-context example, and the task-specific instruction. For brevity, we
truncate the in-context example, and add color styling to help the reader
navigate the text.

\promptheading{Benchmark Instructions}
\begin{PromptBlock}
Interact with a household to solve a task. Imagine you are an intelligent agent
in a household environment and your target is to perform actions to complete
the task goal. At the beginning of your interactions, you will be given the
detailed description of the current environment and your goal to accomplish.

After your each turn, the environment will give you immediate feedback based
on which you plan your next few steps. if the envrionment output
"Nothing happened", that means the previous action is invalid and you should
try more options.
\end{PromptBlock}

\promptheading{Agent Instructions}

We specifically show ReflAct's agent instructions here.

\begin{PromptBlock}
For each of your turn, you will be given the observation of the last turn.
You should first reflect in one sentence on the agent's state in relation to
the task goal, and then output the action for this turn. Your output must
strictly follow this format:\\
"Reflection: your reflection.\\
Action: your next action".

Reminder:\\
1. The action must be chosen from the given available actions. Any actions
except provided available actions will be regarded as illegal.\\
2. You should reflect on the current state in relation to the task goal every
turn before acting.
\end{PromptBlock}

\promptheading{Actions}
\begin{PromptBlock}
The available actions are:\\
1. go to \{recep\}\\
2. take \{obj\} from \{recep\}\\
3. put \{obj\} in/on \{recep\}\\
4. open \{recep\}\\
5. close \{recep\}\\
6. use \{obj\}\\
7. clean \{obj\} with \{recep\}\\
8. heat \{obj\} with \{recep\}\\
9. cool \{obj\} with \{recep\}\\
where \{obj\} and \{recep\} correspond to objects and receptacles.
\end{PromptBlock}

\promptheading{In-Context Example}
\begin{PromptBlock}
---

\promptgap
Here is an example.

\promptgap
You are in the middle of a room. Looking quickly around you, you see a
cabinet 13, a cabinet 12, \textit{[... receptacles omitted ...]}, a
stoveburner 3, a stoveburner 2, a stoveburner 1, and a toaster 1.\\
\tasklabel put a clean lettuce in diningtable.

\promptgap
\reflectionlabel To solve the task, I need to find and take a lettuce, then
clean it with sinkbasin, then put it in diningtable. First I need to find a
lettuce. A lettuce is more likely to appear in fridge 1, diningtable 1,
sinkbasin 1, stoveburner 1-3, cabinet 1-13. I can check one by one, starting
by going to fridge 1.\\
\actionlabel go to fridge 1\\
\observationlabel The fridge 1 is closed.

\promptgap
\reflectionlabel Currently, I am at fridge 1, not holding anything, but the
closed fridge hinders me from finding the lettuce needed to clean and place.\\
\actionlabel open fridge 1\\
\observationlabel You open the fridge 1. The fridge 1 is open. In it, you see
a cup 3, a egg 2, a potato 3, and a potato 2.

\promptgap
\textit{[... intermediate interaction turns omitted ...]}

\promptgap
\reflectionlabel Currently, I am at diningtable 1, still holding the clean
lettuce 1, and the task is nearly complete, with only the placement remaining.\\
\actionlabel put lettuce 1 in/on diningtable 1\\
\observationlabel You put the lettuce 1 in/on the diningtable 1.
\end{PromptBlock}

\promptheading{Task Instruction}
\begin{PromptBlock}
---

\promptgap
Now, it's your turn and here is the task.

You are in the middle of a room. Looking quickly around you, you see a
cabinet 6, a cabinet 5, a cabinet 4, a cabinet 3, a cabinet 2, a cabinet 1,
a coffeemachine 1, a countertop 3, a countertop 2, a countertop 1, a drawer 3,
a drawer 2, a drawer 1, a fridge 1, a garbagecan 1, a microwave 1, a shelf 3,
a shelf 2, a shelf 1, a sinkbasin 1, a stoveburner 4, a stoveburner 3,
a stoveburner 2, a stoveburner 1, and a toaster 1.\\
\tasklabel put a clean mug in coffeemachine.
\end{PromptBlock}

\subsection{World Model Prompt Instructions}
\label{app:alfworld_wm_prompt}

\promptheading{WALL-E World Model Instructions}
\begin{PromptBlock}
A world model checks each action you propose before it is executed. If the
action is infeasible in the current state, the world model will not execute it;
instead it returns an observation beginning with "[World model]" that explains
why the action would fail and suggests how to fix it. This check does not
change the environment and does not consume a step. When you receive a
"[World model]" observation, do not repeat the same action -- read the reason,
address it (for example by first moving to the right location or freeing your
hand), and propose a different action.
\end{PromptBlock}

\promptheading{Belief-Based World Model Instructions}

Below are the instructions used for the \textit{Belief} variant. For BB-WM,
we compose both the WALL-E and these instructions.

\begin{PromptBlock}
You are assisted by a world model that tracks the environment state and where
objects are likely to be. You can consult it for free (a query does not change
the environment) using the query action below.

\promptgap
The available queries are:\\
1. query where is \{obj\}: returns the receptacles where \{obj\} is most
likely to be, ranked by probability (already-searched receptacles are
excluded).\\
2. query what is in \{recep\}: returns the objects observed in \{recep\}, or
tells you it has not been searched yet.\\
3. query searched \{recep\}: tells you whether you have already looked inside
\{recep\}.\\
4. query state: returns a summary of the current state: where you are, what
you are holding, the objects you have seen, and which receptacles you have
searched.

\promptgap
The world model only provides information; it does not tell you what to do.
When searching for an object, you may "query where is \{obj\}" to see which
receptacles are more likely to contain it (already-searched receptacles are
excluded), then decide your own action using that information together with
the task.

\promptgap
Whenever you feel stuck or uncertain about what to do next -- for example, you
cannot find an object, you keep getting "Nothing happened", or you are unsure
of the current state -- query the world model before acting. A query is free
and never changes the environment, so use it to reorient yourself instead of
guessing.
\end{PromptBlock}
\subsection{In-Context Examples}

We use per-task in-context examples, following \citep{xiong-etal-2025-mpo}. They are publicly available in this repo: \href{https://github.com/WeiminXiong/MPO}{https://github.com/WeiminXiong/MPO}

For the \textit{Belief} and \textit{BB-WM} variants, we alter the in-context example to show the agent using the world model to query for the location of the task's target object. Nothing else in the example is changed (following action selection or agentic reasoning thoughts).

\subsection{State Space}
\label{app:alfworld_state}

\subsubsection{Deterministic State Space}
\label{app:alfworld_deterministic_state}

The deterministic component of the world-model state maintains the task goal,
receptacle and object attributes observed so far, and the agent's current
location and inventory. We represent the state at timestep $t$ as
\[
x_t^{\mathrm{det}}
=
\left(
g,\,
\mathcal{R}_t,\,
\mathcal{O}_t,\,
\ell_t,\,
I_t
\right),
\]
with the following components.

\paragraph{Goal.}
The task goal $g$ is fixed for an episode and consists of
\[
g =
(\texttt{act},\,
 \texttt{target\_type},\,
 \texttt{transform},\,
 \texttt{dest\_recep\_type},\,
 \texttt{count}),
\]
where \texttt{act} is one of
\{\texttt{pick\_and\_place}, \texttt{clean}, \texttt{heat},
\texttt{cool}, \texttt{look}, \texttt{picktwo}\}.
The \texttt{transform} and destination fields are optional depending on the
task type.

\paragraph{Receptacles.}
$\mathcal{R}_t$ contains one state entry for each receptacle $r$, with all
receptacle identifiers known at the beginning of the episode:
\[
r =
(\texttt{id},\,
 \texttt{open},\,
 \texttt{searched},\,
 \texttt{contents}).
\]
Here, \texttt{open} $\in \{\texttt{true},\texttt{false},\texttt{unknown}\}$
records the receptacle's open/closed state,
\texttt{searched} indicates whether its contents have been observed, and
\texttt{contents} is the set of observed object identifiers contained in it.

\paragraph{Objects.}
$\mathcal{O}_t$ contains entries for object instances that have been observed
in the environment:
\[
o =
(\texttt{id},\,
 \texttt{type},\,
 \texttt{location},\,
 \texttt{clean},\,
 \texttt{heated},\,
 \texttt{cooled}).
\]
An object's \texttt{location} is a receptacle identifier,
\texttt{inventory}, or \texttt{unknown}. The remaining Boolean attributes
record whether the object has been cleaned, heated, or cooled.

\paragraph{Agent.}
The agent state consists of its current location $\ell_t$ and inventory $I_t$,
where $\ell_t$ is the current receptacle/location and $I_t$ is the set of
object identifiers currently held by the agent.

\subsubsection{Probabilistic State Space}
\label{app:alfworld_probabilistic_state}

The probabilistic component represents uncertainty over the locations of task-relevant objects. It contains only categorical distributions over receptacle locations

Let
\[
\mathcal{R} = \{r_1,\ldots,r_n\}
\]
denote the set of receptacle instances in the scene, all of which are known at
the beginning of the episode. We represent the belief over the target object's
location as
\[
b_t \in \Delta^{|\mathcal{R}|-1},
\]
where
\[
b_t(r) = P(s_t = r)
\]
is the probability that the target object is located at receptacle $r$.
There is no additional ``unknown'' or undiscovered-location category;
probability mass is distributed only over receptacles in $\mathcal{R}$.

\paragraph{Target Belief.}
The belief maintains a categorical distribution over receptacle identifiers,
together with an optional concrete object identifier and observed location:
\[
b_t =
\left(
\texttt{dist},
\texttt{instance\_id},
\texttt{located\_at}
\right).
\]
Before the target object is observed, \texttt{instance\_id} and
\texttt{located\_at} are undefined. Once observed, they identify the concrete
object instance and its known location, and the distribution becomes
degenerate at that receptacle.

Only the task-relevant target object type receives a persistent probabilistic
belief. For tasks requiring multiple objects of the same type, this
representation is extended with a separate target belief for each required
instance.

\paragraph{Prior.}
The initial belief is derived from ALFWorld's object--receptacle placement
constraints. Let
\[
\mathcal{R}_{\mathrm{allow}}(o)
=
\left\{
r \in \mathcal{R}
:
\operatorname{type}(r)
\text{ is a valid receptacle type for } o
\right\}.
\]
The prior is uniform over compatible receptacle instances present in the scene:
\[
b_0(r)
=
\begin{cases}
\dfrac{1}{|\mathcal{R}_{\mathrm{allow}}(o)|},
    & r \in \mathcal{R}_{\mathrm{allow}}(o), \\[6pt]
0,  & \text{otherwise}.
\end{cases}
\]
If no placement information is available for the object type, or no compatible
receptacle type appears in the scene, we instead initialize a uniform
distribution over all $r \in \mathcal{R}$.

\subsection{Belief-Query Interface}
\label{app:alfworld_belief_query_interface}

The policy accesses the world-model state through an explicit query interface.
Queries use the same action channel as environment actions: instead of issuing
an environment action, the agent outputs
\[
\texttt{Action: query <q>}.
\]
The world model answers the query from its current state and returns the result
as the next \texttt{Observation:}. Queries do not modify the environment or
consume an environment step. The interface exposes four query types.

\paragraph{\texttt{query where is \{obj\}}.}
This query exposes the location belief described in
Appendix~\ref{app:alfworld_probabilistic_state}. For the task-relevant target object,
the world model ranks unsearched receptacles with positive belief mass under
$b_t$ and returns the most likely locations and their probabilities, e.g.,
\begin{PromptBlock}
spraybottle is most likely at: cabinet 1 (0.14), cabinet 2 (0.14), ...
\end{PromptBlock}
Locations are returned in decreasing probability order until their cumulative
probability reaches at least $0.9$; any remaining probability mass is
summarized as \texttt{and other receptacles (X\%)}.

For non-target object types, which do not have a persistent belief in the
probabilistic state, the world model instead constructs the corresponding
placement prior over currently unsearched receptacles at query time. If a
specific observed object instance is requested, the response is obtained
directly from the deterministic object state, reporting whether the object is
held, located at a particular receptacle, or has not yet been located.

\paragraph{\texttt{query what is in \{recep\}}.}
This query accesses the deterministic receptacle state. If the receptacle has
been searched, the world model returns its observed contents; otherwise, it
reports that the receptacle has not yet been searched. For example,
\begin{PromptBlock}
cabinet 1 contains: cloth 1, soapbar 1.
\end{PromptBlock}

\paragraph{\texttt{query searched \{recep\}}.}
This query returns whether the specified receptacle has already been searched,
using the \texttt{searched} field of the deterministic state:
\begin{PromptBlock}
cabinet 1: searched
\end{PromptBlock}
or
\begin{PromptBlock}
cabinet 1: not searched yet.
\end{PromptBlock}

\paragraph{\texttt{query state}.}
This query returns a compact summary of the deterministic state described in
Appendix~\ref{app:alfworld_deterministic_state}, including the agent's current
location and inventory, observed objects, and searched receptacles:
\begin{PromptBlock}
You are at {loc}, holding {inv}. Known objects: ... Searched receptacles: ...
\end{PromptBlock}

\section{BB-WM Instantiation for ScienceWorld}

Our ScienceWorld benchmark uses all available environment simplifications, detailed on the repo: \href{https://github.com/allenai/scienceworld}{https://github.com/allenai/scienceworld}

\subsection{General Agent Prompt}
\label{app:scienceworld_agent_prompt}

We provide the agent prompt used for ScienceWorld below. We separate the
prompt into benchmark instructions, agent instructions, the available action
space, the in-context example, and the task-specific instruction. We truncate
the in-context example for brevity and add color styling to help the reader
navigate the text.

\promptheading{Benchmark Instructions}
\begin{PromptBlock}
You are a helpful assistant to do some scientific experiment in an environment.\\
In the environment, there are several rooms: kitchen, foundry, workshop,
bathroom, outside, living room, bedroom, greenhouse, art studio, hallway.\\
You should explore the environment and find the items you need to complete the
experiment.\\
You can teleport to any room in one step.\\
All containers in the environment have already been opened, so you can directly
take items from them.\\
For each of your turn, you will be given the observation of the last turn.
\end{PromptBlock}

\promptheading{Agent Instructions}
\begin{PromptBlock}
You should first reflect on the agent's state, including the location,
inventory, and focused object, in relation to the task goal. Then, output the
action for this turn. Your output must strictly follow this format:\\
"Reflection: your reflection.\\
Action: your next action".
\end{PromptBlock}

\promptheading{Actions}
\begin{PromptBlock}
The available actions are:\\
open OBJ: open a container\\
close OBJ: close a container\\
activate OBJ: activate a device\\
deactivate OBJ: deactivate a device\\
connect OBJ to OBJ: connect electrical components\\
disconnect OBJ: disconnect electrical components\\
use OBJ [on OBJ]: use a device/item\\
look around: describe the current room\\
examine OBJ: describe an object in detail\\
look at OBJ: describe a container's contents\\
read OBJ: read a note or book\\
move OBJ to OBJ: move an object to a container\\
pick up OBJ: move an object to the inventory\\
pour OBJ into OBJ: pour a liquid into a container\\
mix OBJ: chemically mix a container\\
teleport to LOC: teleport to a specific room\\
focus on OBJ: signal intent on a task object\\
wait: take no action for 10 steps\\
wait1: take no action for a step
\end{PromptBlock}

\promptheading{In-Context Example}
\begin{PromptBlock}
---\\
Here is an example.

\promptgap
Task Description:\\
\tasklabel boil orange juice. For compounds without a boiling point,
combusting the substance is also acceptable. First, focus on the substance.
Then, take actions that will cause it to change its state of matter.

\promptgap
\reflectionlabel The task is to boil the orange juice, which means heating it
until it changes state. I will need a container, a heat source, and a
thermometer. The kitchen has these, so I teleport there.\\
\actionlabel teleport to kitchen\\
\observationlabel You teleport to the kitchen.

\promptgap
\reflectionlabel I look around the kitchen to find the thermometer, a pot, and
the orange juice.\\
\actionlabel look around\\
\observationlabel This room is called the kitchen. In it, you see:\\
\hspace*{1em}the agent\\
\hspace*{1em}a substance called air\\
\hspace*{1em}a chair. On the chair is: nothing.\\
\hspace*{1em}a counter. On the counter is: a bowl (containing a red apple, a
banana, an orange, a potato), a drawer.\\
\hspace*{1em}a cupboard. The cupboard door is open. In the cupboard is: a
ceramic cup (containing nothing), a drawer, a metal pot (containing nothing),
a tin cup (containing nothing).\\
\hspace*{1em}a freezer. The freezer door is open. In the freezer is: nothing.\\
\hspace*{1em}a fridge. The fridge door is open. In the fridge is: a wood cup
(containing orange juice).\\
\hspace*{1em}a glass jar (containing a substance called sodium chloride)\\
\hspace*{1em}a lighter\\
\hspace*{1em}an oven, which is turned off. The oven door is open. In the oven
is: nothing.\\
\hspace*{1em}a painting\\
\hspace*{1em}a sink, which is turned off. In the sink is: nothing.\\
\hspace*{1em}a substance called soap\\
\hspace*{1em}a stopwatch, which is deactivated.\\
\hspace*{1em}a stove, which is turned off. On the stove is: nothing.\\
\hspace*{1em}a table. On the table is: a glass cup (containing nothing).\\
\hspace*{1em}a thermometer, currently reading a temperature of 10 degrees
celsius\\
You also see:\\
\hspace*{1em}A door to the bathroom (that is open)\\
\hspace*{1em}A door to the hallway (that is open)\\
\hspace*{1em}A door to the outside (that is open)

\promptgap
\reflectionlabel I pick up the thermometer so I can monitor the temperature as
I heat the juice.\\
\actionlabel pick up thermometer\\
\observationlabel You move the thermometer to the inventory.

\promptgap
\textit{[... intermediate interaction turns omitted ...]}

\promptgap
\reflectionlabel I turn the blast furnace on to heat the juice toward
boiling.\\
\actionlabel activate blast furnace\\
\observationlabel The blast furnace is now activated.

\promptgap
\reflectionlabel I use the thermometer on the orange juice to check its
temperature as it heats.\\
\actionlabel use thermometer in inventory on orange juice\\
\observationlabel the thermometer measures a temperature of 24 degrees celsius

\promptgap
\reflectionlabel The juice has boiled into gas; I measure the gaseous juice to
confirm it changed state.\\
\actionlabel use thermometer in inventory on gaseous orange juice\\
\observationlabel No known action matches that input.

\promptgap
\reflectionlabel I wait to let the blast furnace finish boiling it off.\\
\actionlabel wait1\\
\observationlabel You decide to wait for 1 iterations.
\end{PromptBlock}

\promptheading{Task Instruction}
\begin{PromptBlock}
---

\promptgap
Now, it's your turn and here is the task.

\promptgap
Task Description:\\
\tasklabel boil lead. For compounds without a boiling point, combusting the
substance is also acceptable. First, focus on the substance. Then, take actions
that will cause it to change its state of matter.
\end{PromptBlock}

\subsection{World Model Prompt Instructions}
\label{app:scienceworld_wm_prompt}

\promptheading{WALL-E Oracle World Model Instructions}
\begin{PromptBlock}
A world model checks each action before it runs. If you propose an action that
is not possible in the current state, it will not be executed; instead you will
receive an observation beginning with "[World model]" explaining that the
action is unavailable. When you see such a message, do not repeat the same
action -- choose a different, valid action.
\end{PromptBlock}

\promptheading{Belief-Based World Model Instructions}
\begin{PromptBlock}
You are assisted by a world model that tracks where objects are likely to be.
You may consult it at any time with a query action instead of an environment
action; a query does not consume an environment step.

\promptgap
You may also consult the world model (these do not consume an environment
step):\\
query where is OBJ: ask the world model which room(s) an object is most likely
in\\
query what is in ROOM: ask what the world model has seen in a room you have
visited\\
query searched ROOM: ask whether you have already looked around a room\\
query state: ask for your current location, inventory, focused object, and
searched rooms

\promptgap
Reminder:\\
1. The world model only provides information; it does not tell you what to do.
For example, when searching for an object, you may "query where is \{obj\}" to
see which rooms are more likely to contain it (already-searched rooms are
excluded), then decide your own action using that information together with the
task.\\
2. Whenever you feel stuck or uncertain about what to do next -- for example,
you cannot find an object, you keep getting "No known action matches that
input", or you are unsure of the current state -- query the world model before
acting. A query is free and never changes the environment, so use it to
reorient yourself instead of guessing.
\end{PromptBlock}

\subsection{In-Context Examples}

Following our process for ALFWorld, we create per-task in-context examples. We generate in-context examples by using the environments \textit{gold path generator}. We distill the gold path action sequence by dropping redundant actions, and we verify on a fresh run that the distilled action sequence succeeds in solving the task. Reasoning thoughts (for use in ReAct and ReflAct) frameworks are written by an agent.

\subsection{State Space}

\subsubsection{Deterministic State Space}
\label{app:scienceworld_deterministic_state}

The deterministic component of the ScienceWorld state maintains the
task goal, room contents, object nesting, and the agent's current location,
inventory, and focused object. All quantities in this component represent
observed or task-specified facts. We represent the deterministic state at timestep $t$ as
\[
x_t^{\mathrm{det}}
=
\left(
g,\,
\mathcal{R}_t,\,
\mathcal{C}_t,\,
\mathcal{L}_t,\,
\ell_t,\,
I_t,\,
f_t
\right),
\]
with the following components.

\paragraph{Goal.}
The task goal $g$ is fixed for an episode and contains
\[
g =
\left(
\texttt{target\_type},\,
\texttt{named\_location}
\right),
\]
where \texttt{target\_type} identifies the task-relevant object (e.g.,
\texttt{orange juice} or \texttt{aluminum foil}), and
\texttt{named\_location} optionally specifies a room in which the task
description explicitly states that the target is located.

\paragraph{Rooms.}
ScienceWorld uses a fixed set of ten rooms,
\[
\mathcal{R}
=
\left\{
\begin{gathered}
\texttt{kitchen},
\texttt{bathroom},
\texttt{living room},
\texttt{bedroom},
\texttt{workshop}, \\
\texttt{greenhouse},
\texttt{art studio},
\texttt{foundry},
\texttt{outside},
\texttt{hallway}
\end{gathered}
\right\}.
\]

For each room $r$, the deterministic state maintains
\[
r =
\left(
\texttt{name},\,
\texttt{searched},\,
\texttt{contents}
\right),
\]
where \texttt{searched} indicates whether the room has been observed via the
\texttt{look around} action, and \texttt{contents} is the set of object referents observed in that room.

Unlike ALFWorld, ScienceWorld objects are represented directly by potentially
multi-word referents such as \texttt{orange juice}, \texttt{red apple}, or
\texttt{metal pot}, rather than numbered object identifiers.

\paragraph{Object Nesting.}
We record observed containment relationships through
\[
\mathcal{C}_t :
c \mapsto \{o_1,\ldots,o_m\},
\]
where $\mathcal{C}_t(c)$ is the set of objects observed inside or on container
$c$. An inverse object-location map
\[
\mathcal{L}_t :
o \mapsto (r,c)
\]
records the room and, when applicable, container associated with an observed
object.

For held containers, the state additionally maintains their known contents:
\[
\mathcal{H}_t :
c \mapsto
\begin{cases}
\{o_1,\ldots,o_m\}, & \text{known contents},\\
\emptyset, & \text{known to be empty},\\
\texttt{unknown}, & \text{contents not known}.
\end{cases}
\]

\paragraph{Agent.}
The agent state consists of its current room $\ell_t$, inventory $I_t$, and
focused object $f_t$:
\[
a_t^{\mathrm{state}}
=
\left(
\ell_t,\,
I_t,\,
f_t
\right).
\]
Here, $\ell_t$ is the agent's current room, $I_t$ is the set of object
referents currently held, and $f_t$ is the object most recently selected by a
\texttt{focus on} action.

\subsubsection{Probabilistic State Space}
\label{app:scienceworld_probabilistic_state}

The probabilistic component of the ScienceWorld world model represents
uncertainty over the room containing an object. Let $\mathcal{R}$ denote the fixed set of ten ScienceWorld rooms. For a tracked
object type $o$, we maintain a categorical belief
\[
b_t^o \in \Delta^{|\mathcal{R}|-1},
\]
where
\[
b_t^o(r)
=
P(s_t^o = r),
\qquad r \in \mathcal{R},
\]
is the probability that an instance of object type $o$ is located in room $r$.

Each object belief additionally records a resolved location,
\[
b_t^o =
\left(
\texttt{dist},
\texttt{located\_at}
\right),
\]
where \texttt{dist} is the categorical distribution over rooms and
\texttt{located\_at} is initially undefined. Once the object is observed,
\texttt{located\_at} records its room; if the object is held by the agent, it
takes the value \texttt{inventory}. The target object specified by the task is
tracked from the beginning of the episode. Additional object types may be
instantiated as needed and subsequently follow the same belief representation
and update rules.

\paragraph{Prior.}
The initial belief is derived from ScienceWorld's object--room placement constraints. Let
\[
\mathcal{R}_{\mathrm{allow}}(o)
\subseteq \mathcal{R}
\]
denote the set of rooms in which object type $o$ may occur. In the absence of
additional task information, the prior is uniform over these candidate rooms:
\[
b_0^o(r)
=
\begin{cases}
\dfrac{1}{|\mathcal{R}_{\mathrm{allow}}(o)|},
    & r \in \mathcal{R}_{\mathrm{allow}}(o), \\[6pt]
0,  & \text{otherwise}.
\end{cases}
\]
If no placement information is available for an object type, the prior is
uniform over all rooms in $\mathcal{R}$.

Task descriptions may provide additional location information. If the task
explicitly specifies the target object's room $r^\star$, the target belief is
initialized deterministically:
\[
b_0^o(r)
=
\mathbbm{1}[r=r^\star].
\]
For other objects whose locations are stated in the task description, we place
$0.9$ probability on the stated room and distribute the remaining $0.1$
uniformly over the object's other candidate rooms.

\subsection{Belief-Query Interface}
\label{app:scienceworld_belief_query_interface}

The policy accesses the ScienceWorld world-model state through an explicit
query interface. Queries use the same action channel as environment actions:
instead of issuing an environment action, the agent outputs
\[
\texttt{Action: query <q>}.
\]
The world model answers from its current state and returns the result as the
next \texttt{Observation:}. Queries do not modify the ScienceWorld environment
or consume an environment step. The interface exposes four query types.

\paragraph{\texttt{query where is OBJ}.}
This query exposes the room-level location belief described in
Appendix~\ref{app:scienceworld_probabilistic_state}. For an unresolved object
with a maintained belief $b_t^o$, the world model ranks rooms with positive
belief mass that have not yet been searched and returns the most likely
locations and their probabilities, e.g.,
\begin{PromptBlock}
orange juice is most likely in: kitchen (1.00).
\end{PromptBlock}
Rooms are returned in decreasing probability order until their cumulative
probability reaches at least $0.9$; any remaining mass is summarized as
\texttt{and other rooms (X\%)}.

Once an object has been observed, the query may instead resolve its location
from the deterministic state described in
Appendix~\ref{app:scienceworld_deterministic_state}. In particular, if the
object has been observed inside a container, the response includes both the
room and container:
\begin{PromptBlock}
orange juice is in the kitchen, in the fridge.
\end{PromptBlock}
If the object is held by the agent, its location is reported as
\texttt{inventory} with probability $1$. If all candidate rooms have been
searched without finding the object, the world model reports that no candidate
rooms remain.

\paragraph{\texttt{query what is in X}.}
This query accesses the deterministic room and container contents maintained by
the world model. If $X$ is a searched room, the query returns the objects
observed in that room. If $X$ is an observed container, it instead returns the
objects recorded inside that container. For example,
\begin{PromptBlock}
fridge contains: orange juice.
\end{PromptBlock}
If the requested room or container has not yet been observed, the world model
reports that it has not been searched.

\paragraph{\texttt{query searched ROOM}.}
This query reports whether a \texttt{look around} observation has been obtained
for the specified room, corresponding to the \texttt{searched} field in the
deterministic room state:
\begin{PromptBlock}
kitchen: searched
\end{PromptBlock}
or
\begin{PromptBlock}
kitchen: not searched yet.
\end{PromptBlock}

\paragraph{\texttt{query state}.}
This query returns a compact summary of the deterministic agent state described
in Appendix~\ref{app:scienceworld_deterministic_state}, including the current
room, inventory, focused object, and searched rooms:
\begin{PromptBlock}
You are in the {loc}, holding {inv}, focused on {focus}. Rooms searched: ...
\end{PromptBlock}
When the agent holds a container, its known contents are included in the
inventory summary, e.g., \texttt{metal pot (containing orange juice)}.

\section{BB-WM Instantiation for BabyAI}
\label{app:babyai}

We evaluate on the following task families from BALROG: \texttt{goto}, \texttt{pickup}, \texttt{pick\_up\_seq\_go\_to}, \texttt{putnext}. We omit \texttt{open}, as it requires a different environment configuration than the other four tasks.

\subsection{General Agent Prompt}
\label{app:babyai_agent_prompt}

We provide the agent prompt used for BabyAI below. We separate the prompt
into benchmark instructions, agent instructions, the available action space,
tips, and the task-specific instruction. There is no in-context example.

\promptheading{Benchmark Instructions}
\begin{PromptBlock}
You are an agent playing a simple navigation game. You are in a partially
observable grid environment. You can only observe objects in your current
field of view.
\end{PromptBlock}

\promptheading{Agent Instructions}

We specifically show ReflAct's agent instructions here.

\begin{PromptBlock}
For each of your turn, you will be given the observation of the last turn.
You should first reflect in one sentence on the agent's state in relation to
the task goal, and then output the action for this turn. Your output must
strictly follow this format:\\
"Reflection: your reflection.\textbackslash n Action: your next action".

\promptgap
Remember that you can only output one "Action:" in per response. You should
reflect on the current state in relation to the task goal every turn before
acting.
\end{PromptBlock}

\promptheading{Actions}
\begin{PromptBlock}
The following are the possible actions you can take in the game, followed by
a short description of each action:\\
turn left: turn to the left,\\
turn right: turn to the right,\\
go forward: take one step forward,\\
pick up: pick up the object directly in front of you,\\
drop: drop the object that you are holding,\\
toggle: manipulate the object in front of you.
\end{PromptBlock}

\promptheading{Tips}
\begin{PromptBlock}
Tips:\\
- Once the desired object you want to interact or pickup is in front of you,
you can use the 'toggle' action to interact with it (or 'pick up' to take it).\\
- It doesn't make sense to repeat the same action over and over if the
observation doesn't change.\\
- Any action except the six listed above (and the query actions, if provided)
is illegal and will not change the environment.
\end{PromptBlock}

\promptheading{Task Instruction}
\begin{PromptBlock}
Now, it's your turn and here is the task.\\
\tasklabel go to the green key.\\
a wall 3 steps forward\\
a wall 4 steps left\\
a red ball 1 step forward\\
a purple box 2 steps forward and 1 step right
\end{PromptBlock}

\subsection{World Model Prompt Instructions}
\label{app:babyai_wm_prompt}

\promptheading{WALL-E World Model Instructions}
\begin{PromptBlock}
A world model checks each action you propose before it is executed. If the
action is infeasible in the current state, the world model will not execute it;
instead it returns an observation beginning with "[World model]" that explains
why the action would fail and suggests how to fix it. This check does not
change the environment and does not consume a step. When you receive such an
observation, do not repeat the same action -- read the reason, address it (for
example by first moving to the right location or freeing your hand), and
propose a different action.
\end{PromptBlock}

When WALL-E is used alone, the prompt also includes the following tip:
\begin{PromptBlock}
- The world model only provides information; it does not tell you what to do.
\end{PromptBlock}

\promptheading{Belief-Based World Model Instructions}

Below are the instructions used for the \textit{Belief} variant. For BB-WM,
we compose both the WALL-E and these instructions. In that composition the
belief paragraph begins ``The same world model also tracks a 6x6 room prior
and where unseen objects are likely to be.''

\begin{PromptBlock}
You are assisted by a model that tracks a 6x6 room prior and where unseen
objects are likely to be. You can consult it for free (a query does not
change the environment) using the query action below:\\
1. query where is \{obj\}: if seen, last-known place; if not, remaining
candidate cells (nearest first) once the room is localized\\
2. query map: print the pinned 6x6 room (? = still possible, . = ruled out,
letter = seen object)\\
3. query what is at \{place\}: ask what the world model has recorded at an
egocentric place (e.g. "1 step forward")\\
4. query searched \{place\}: ask whether that place has been observed\\
5. query what have I seen: list objects the world model has recorded\\
6. query state: the 6x6 map and whether you are holding anything

\promptgap
- The world model only provides information; it does not tell you what to do.\\
- Whenever you feel stuck or uncertain about what to do next -- for example,
you cannot find an object or you are unsure of the current state -- query the
world model before acting. A query is free and never changes the environment,
so use it to reorient yourself instead of guessing.
\end{PromptBlock}

\subsection{State Space}
\label{app:babyai_state}

\subsubsection{Deterministic State Space}
\label{app:babyai_deterministic_state}

The deterministic component is a partial map built from the text observations.
The world model does not use MiniGrid's coordinates. It keeps its own grid,
fixed at the start of the episode: the cell where the agent begins is
$(0,0)$, and the direction the agent faces at initialization is $+Y$
(\texttt{start-forward}). Later turns and steps are recorded in that same
grid. An observation such as ``a red ball 1 step forward'' is converted from
the agent's current position and heading into one cell of this grid.

The map starts with only that origin cell. A cell is added when an
observation names it, or when a successful \texttt{go forward} moves the
agent onto it. A line ``a wall $N$ steps forward'' adds a wall $N$ cells
along the current heading and marks the cells in between as empty. An object
line adds that object at the named offset. Cells that have neither been
described nor stepped on are left out of the map. We represent the state at timestep $t$ as
\[
x_t^{\mathrm{det}}
=
\left(
g,\,
\mathcal{C}_t,\,
\mathcal{O}_t,\,
\ell_t,\,
I_t
\right),
\]
with the following components.

\paragraph{Goal.}
The mission $g$ is fixed for an episode and consists of
\[
g =
(\texttt{act},\,
 \texttt{referents},\,
 \texttt{seq\_order}),
\]
where \texttt{act} is one of
\{\texttt{goto}, \texttt{pickup}, \texttt{putnext}, \texttt{seq}\}.
Each referent is a color together with a type in
\{\texttt{key}, \texttt{ball}, \texttt{box}\}.
\texttt{seq\_order} is \texttt{then} or \texttt{after}
for pick-up-then-go-to missions, and is otherwise absent.

\paragraph{Pose.}
The agent pose in this grid is
\[
\ell_t = (x_t,\, y_t,\, h_t),
\]
with heading $h_t \in \{0,1,2,3\}$ named
\texttt{start-forward}, \texttt{start-right}, \texttt{start-back}, and
\texttt{start-left}, in clockwise order from the initial facing.
\texttt{turn left} and \texttt{turn right} update $h_t$ directly.
\texttt{go forward} updates $(x_t, y_t)$ only when the new observation
differs from the previous one and the cell in front is not already known
to be blocked. An unchanged observation is treated as a failed step, so the
agent stays in place.

\paragraph{Cells.}
$\mathcal{C}_t$ maps coordinates in this grid to the cells recorded so far.
A cell $c$ has
\[
c =
(\texttt{kind},\,
 \texttt{color},\,
 \texttt{type},\,
 \texttt{last\_seen\_step},\,
 \texttt{visited}),
\]
where \texttt{kind} is one of
\{\texttt{empty}, \texttt{wall}, \texttt{object}\}.
\texttt{visited} is true for cells the agent has occupied, including the
origin.

\paragraph{Objects.}
$\mathcal{O}_t$ is indexed by color and type, such as \texttt{green key}; there are no instance id's. It records every object that appears in the observations, and it's location (a cell in the grid or inventory).

\paragraph{Inventory.}
$I_t$ is the carried object, a color--type pair parsed from observations (e.g., ``You carry
\dots''), or empty.

\subsubsection{Probabilistic State Space}
\label{app:babyai_probabilistic_state}

The probabilistic component is a uniform placement prior over BabyAI's $6\times 6$ walkable room. Since the starting orientation of the agent is not known, the room cannot be localized, or \textit{pinned}, until at least two adjacent walls are observed. 

Let $r$ be a mission referent. We represent its location belief as
\[
b_t^{(r)} \in \Delta^{|\mathcal{A}^{(r)}|-1},
\qquad
b_t^{(r)}(a) = P(s_t^{(r)} = a),
\]
where the atoms $\mathcal{A}^{(r)}$ are \texttt{UNSEEN} before localization
and the remaining candidate cells afterward. A sequential mission keeps one
belief for each referent. Referents that share a color and type share one
belief. Objects that are not in the mission have no persistent belief;
queries about them use $\mathcal{O}_t$.

\paragraph{Target Belief.}
Each belief stores the categorical distributiontogether with an optional located atom:
\[
b_t^{(r)}
=
\left(
\texttt{dist},\,
\texttt{located\_at}
\right).
\]
Before the referent is observed, \texttt{located\_at} is undefined. After being observed, location becomes that cell, and on pickup it becomes \texttt{inventory}.

\paragraph{Prior.}
Before the room is localized, the belief is a point mass on the unseen atom,
\[
b_0^{(r)}(\texttt{UNSEEN}) = 1.
\]
Once the room is pinned, that mass is replaced by a uniform distribution over
interior cells the generator can still occupy. Let $\mathcal{C}_{\mathrm{allow}}(r)$
be the interior cells that \textit{are not} the start cell, every cell within Manhattan
distance $2$ of the start, the agent's current cell, and any cell already
known to be empty, a wall, or occupied by a different object. Then
\[
b(c)
=
\begin{cases}
\dfrac{1}{|\mathcal{C}_{\mathrm{allow}}(r)|},
    & c \in \mathcal{C}_{\mathrm{allow}}(r), \\[6pt]
0,  & \text{otherwise}.
\end{cases}
\]
Later observations remove ruled-out cells and renormalize. The first sighting replaces the prior with a point mass, as above.

\subsection{Belief-Query Interface}
\label{app:babyai_belief_query_interface}

The policy accesses the world-model state through an explicit query interface.
Queries use the same action channel as environment actions: instead of issuing
an environment action, the agent outputs
\[
\texttt{Action: query <q>}.
\]
The world model answers from its current state and returns the result as the
next \texttt{Observation:}. Queries do not modify the environment or consume
an environment step. Responses are egocentric (``1 step forward'', ``2 steps
back and 1 step right'', ``front''). The interface exposes six query types.

\paragraph{\texttt{query where is \{obj\}}.}
If the object is carried, the answer is taken from $I_t$. If it has been
observed, the answer is the last-known atom in $\mathcal{O}_t$, rendered
relative to the current pose, e.g.,
\begin{PromptBlock}
The green key is at 2 steps forward and 1 step right.
\end{PromptBlock}
If it has not been seen and the room is not yet localized, the world model
asks for a wall ahead and a wall to one side:
\begin{PromptBlock}
The green key has not been seen. The 6x6 room is not localized yet
(observe a wall in front and a wall to the left or right).
\end{PromptBlock}
Once the room is localized, the answer is the placement prior of
Appendix~\ref{app:babyai_probabilistic_state}: the number of remaining cells,
the nearest cells, and which side of the agent still holds mass, e.g.,
\begin{PromptBlock}
The green key has not been seen. 12 candidate cells remain (about 8\% each).
Nearest: 2 steps forward, 2 steps forward and 1 step right, and 10 farther
cells. Mass is in front of you (40\%), to your right (35\%).
\end{PromptBlock}

\paragraph{\texttt{query map}.}
This query prints the pinned room. "\texttt{?}" marks a cell that still has
prior mass, "\texttt{.}" a cell that has been ruled out, and a letter marks a seen
object. Before both wall axes are known, the answer states that the room is
not localized yet.

\paragraph{\texttt{query what is at \{place\}}.}
This query reads $\mathcal{C}_t$ at the cell corresponding to the egocentric
place. If that cell has been written, the world model returns its fact;
otherwise it reports whether the cell is still a candidate, ruled out, or
not yet observed. For example,
\begin{PromptBlock}
1 step forward is a red ball.
\end{PromptBlock}

\paragraph{\texttt{query searched \{place\}}.}
A place counts as searched when its cell is present in $\mathcal{C}_t$:
\begin{PromptBlock}
1 step forward: searched.
\end{PromptBlock}
or
\begin{PromptBlock}
2 steps left: not searched yet.
\end{PromptBlock}

\paragraph{\texttt{query what have I seen}.}
This query lists referent keys in $\mathcal{O}_t$ with their last-known
places, e.g.,
\begin{PromptBlock}
You have seen: red ball at 1 step forward, green key at 2 steps forward.
\end{PromptBlock}

\paragraph{\texttt{query state}.}
This query returns the inventory and the same map as \texttt{query map}:

\subsection{WALL-E Implementation}
\label{app:babyai_walle}

Our hand-crafted implementation of WALL-E checks a proposed action against the environment's state before it is executed. If the action would fail, WALL-E
does not step the environment. The agent instead receives some feedback and asked to reconsider the action. Any case not handled below is left to the environment, which if invalid, returns \texttt{Nothing happens. That is not a valid action.} and does count as a step.

\paragraph{\texttt{go forward}.}
The step is rejected when the cell in front blocks movement. On these tasks
that cell is a wall or an object.
\begin{PromptBlock}
[World model] You cannot go forward because there is a wall in the way.
\end{PromptBlock}
\begin{PromptBlock}
[World model] You cannot go forward because there is a red ball in the way.
\end{PromptBlock}

\paragraph{\texttt{pick up}.}
Pickup uses the cell in front of the agent. It is rejected when that cell is
empty, when it cannot be carried, or when the agent is already holding
something.
\begin{PromptBlock}
[World model] You cannot pick up anything because there is no object in front of you.
\end{PromptBlock}
\begin{PromptBlock}
[World model] You cannot pick up the grey wall.
\end{PromptBlock}
\begin{PromptBlock}
[World model] You cannot pick up the red ball because you are already carrying a grey key.
\end{PromptBlock}

\paragraph{\texttt{drop}.}
Drop is rejected when the agent is holding nothing, or when the cell in front
is already occupied.
\begin{PromptBlock}
[World model] You cannot drop anything because you are not holding an object.
\end{PromptBlock}
\begin{PromptBlock}
[World model] You cannot drop the grey key because there is a wall in the way.
\end{PromptBlock}

\paragraph{\texttt{toggle}.}
Toggle is rejected only when the cell in front is empty. A toggle aimed at a
wall or an object is executed.
\begin{PromptBlock}
[World model] You cannot toggle anything because there is no object in front of you.
\end{PromptBlock}

\paragraph{Movement aliases.}
\texttt{go back}, \texttt{go backward}, and \texttt{go left} or
\texttt{go right} (including \texttt{move}, \texttt{walk}, and \texttt{step},
with or without a step count) are not MiniGrid actions. WALL-E rejects them
and names the legal verbs. The same template is used for \texttt{left} and
\texttt{right}, with that direction in place of \texttt{backward}.
\begin{PromptBlock}
[World model] You cannot go backward. Valid actions are: turn left, turn right, go forward, pick up, drop, toggle. To go backward, you must turn to face that direction, then go forward.
\end{PromptBlock}

\section{Example ALFWorld Trajectories}
\label{app:alfworld_trajectory}

We provide an example ALFWorld trajectory (task \#10 from the unseen split) from Sonnet+ReAct using WALL-E and BB-WM. For readability, we color-code agent thoughts, actions, and environment observations, and we \hl{highlight in yellow} critical parts of the trajectory.

\promptheading{WALL-E}

We provide an example ALFWorld trajectory (task \#10 from the unseen split) from Sonnet+ReAct using WALL-E. For readability, we color-code agent thoughts, actions, and environment observations, and we \hl{highlight in yellow} critical parts of the trajectory, which show why the agent misses searching the \texttt{garbagecan} for the egg.

\begin{PromptBlock}
You are in the middle of a room. Looking quickly around you, you see a
cabinet 6, a cabinet 5, a cabinet 4, a cabinet 3, a cabinet 2, a cabinet 1,
a coffeemachine 1, a countertop 3, a countertop 2, a countertop 1, a drawer 3,
a drawer 2, a drawer 1, a fridge 1, a garbagecan 1, a microwave 1, a shelf 3,
a shelf 2, a shelf 1, a sinkbasin 1, a stoveburner 4, a stoveburner 3,
a stoveburner 2, a stoveburner 1, and a toaster 1.

\promptgap
\tasklabel heat some egg and put it in garbagecan.

\promptgap
\thoughtlabel I need to find an egg, heat it with the microwave, then put it
in the garbagecan. Let me start by \hlreason{looking for an egg in likely
locations}.\\
\actionlabel go to fridge 1\\
\observationlabel The fridge 1 is closed.

\promptgap
\actionlabel open fridge 1\\
\observationlabel You open the fridge 1. The fridge 1 is open. In it, you see
a apple 1, a cup 2, a cup 1, a lettuce 2, a potato 2, and a potato 1.

\promptgap
\thoughtlabel No egg in the fridge. \hlreason{Let me check other locations like
countertops, shelves, and cabinets}.\\
\actionlabel go to countertop 1\\
\observationlabel On the countertop 1, you see a cellphone 3, a pan 1, and a
peppershaker 2.

\promptgap
\textit{[... checking other countertop locations ...]}

\promptgap
\thoughtlabel No egg on countertops. \hlreason{Let me check shelves and
cabinets}.\\
\actionlabel go to shelf 1\\
\observationlabel On the shelf 1, you see a saltshaker 2.

\promptgap
\textit{[... checking other shelf and cabinet locations ...]}

\promptgap
\thoughtlabel No egg found in cabinets. Let me check drawers.\\
\actionlabel go to drawer 1\\
\observationlabel The drawer 1 is closed.

\promptgap
\textit{[... checking other drawer locations ...]}

\promptgap
\thoughtlabel No egg found in drawers either. Let me check the stoveburners
and other locations I might have missed.\\
\actionlabel go to stoveburner 1\\
\observationlabel On the stoveburner 1, you see a pan 1.

\promptgap
\textit{[... checking other stoveburner locations ...]}

\promptgap
\actionlabel go to stoveburner 4\\
\observationlabel On the stoveburner 4, you see a pot 1.

\promptgap
\textbf{Outcome:} Failed\\
\textbf{Environment steps:} 30\\
\textbf{WALL-E rejections:} 0
\end{PromptBlock}

The trajectory terminates because the agent hits the task's action budget. 

\promptheading{BB-WM}
\begin{PromptBlock}
You are in the middle of a room. Looking quickly around you, you see a
cabinet 6, a cabinet 5, a cabinet 4, a cabinet 3, a cabinet 2, a cabinet 1,
a coffeemachine 1, a countertop 3, a countertop 2, a countertop 1, a drawer 3,
a drawer 2, a drawer 1, a fridge 1, a garbagecan 1, a microwave 1, a shelf 3,
a shelf 2, a shelf 1, a sinkbasin 1, a stoveburner 4, a stoveburner 3,
a stoveburner 2, a stoveburner 1, and a toaster 1.

\promptgap
\tasklabel heat some egg and put it in garbagecan.

\promptgap
\thoughtlabel I need to find an egg, heat it with the microwave, then put it
in the garbagecan. Let me first find the egg.\\
\actionlabel \hlreason{query where is egg}\\
\observationlabel egg is most likely at: countertop 1 (0.14), countertop 2
(0.14), countertop 3 (0.14), fridge 1 (0.14), garbagecan 1 (0.14),
microwave 1 (0.14), sinkbasin 1 (0.14).

\promptgap
\actionlabel go to fridge 1\\
\observationlabel The fridge 1 is closed.

\promptgap
\actionlabel open fridge 1\\
\observationlabel You open the fridge 1. The fridge 1 is open. In it, you see
a apple 1, a cup 2, a cup 1, a lettuce 2, a potato 2, and a potato 1.

\promptgap
\actionlabel go to countertop 1\\
\observationlabel On the countertop 1, you see a cellphone 3, a pan 1, and a
peppershaker 2.

\promptgap
\textit{[... checking other countertop locations ...]}

\promptgap
\thoughtlabel No egg on the countertops.
\hlreason{Let me check microwave and other locations.}\\
\actionlabel go to microwave 1\\
\observationlabel The microwave 1 is closed.

\promptgap
\actionlabel open microwave 1\\
\observationlabel You open the microwave 1. The microwave 1 is open. In it,
you see nothing.

\promptgap
\actionlabel \hlreason{query where is egg}\\
\observationlabel egg is most likely at: garbagecan 1 (0.50), sinkbasin 1
(0.50).

\promptgap
\actionlabel go to sinkbasin 1\\
\observationlabel On the sinkbasin 1, you see a fork 2, a spatula 2, and a
spoon 3.

\promptgap
\actionlabel go to garbagecan 1\\
\observationlabel On the garbagecan 1, you see a apple 2, a egg 2, and a
egg 1.

\promptgap
\actionlabel take egg 1 from garbagecan 1\\
\observationlabel You pick up the egg 1 from the garbagecan 1.

\promptgap
\actionlabel heat egg 1 with microwave 1\\
\observationlabel \hlreason{[World model] Failed to heat egg 1. It must be
placed in microwave 1 to be heated. Place egg 1 in microwave 1 before heating.}

\promptgap
\actionlabel go to microwave 1\\
\observationlabel The microwave 1 is open. In it, you see nothing.

\promptgap
\actionlabel put egg 1 in/on microwave 1\\
\observationlabel You put the egg 1 in/on the microwave 1.

\promptgap
\actionlabel heat egg 1 with microwave 1\\
\observationlabel Nothing happens.

\promptgap
\thoughtlabel Let me try closing the microwave first and then heating.\\
\actionlabel close microwave 1\\
\observationlabel You close the microwave 1.

\promptgap
\actionlabel heat egg 1 with microwave 1\\
\observationlabel Nothing happens.

\promptgap
\thoughtlabel
\hlreason{Let me try opening the microwave and taking the egg out, then using
the heat action differently.}\\
\actionlabel open microwave 1\\
\observationlabel You open the microwave 1. The microwave 1 is open. In it,
you see a egg 1.

\promptgap
\actionlabel take egg 1 from microwave 1\\
\observationlabel You pick up the egg 1 from the microwave 1.

\promptgap
\actionlabel heat egg 1 with microwave 1\\
\observationlabel You heat the egg 1 using the microwave 1.

\promptgap
\actionlabel go to garbagecan 1\\
\observationlabel On the garbagecan 1, you see a apple 2, and a egg 2.

\promptgap
\actionlabel put egg 1 in/on garbagecan 1\\
\observationlabel You put the egg 1 in/on the garbagecan 1.

\promptgap
\textbf{Outcome:} Success\\
\textbf{Environment steps:} 20\\
\textbf{Belief query steps:} 2\\
\textbf{WALL-E rejections:} 1
\end{PromptBlock}

\section{Example ScienceWorld Trajectories}
\label{app:scienceworld_trajectory}

We provide an example ScienceWorld trajectory (grow plant task, \#93 from the unseen split) from Qwen3-14B+ReflAct using WALL-E Oracle and BB-WM. For readability, we color-code agent thoughts, actions, and environment observations, and we \hl{highlight in yellow} critical parts of the trajectory.

\promptheading{WALL-E Oracle}
\begin{PromptBlock}
Task Description:\\
\tasklabel grow a orange plant from seed. Seeds can be found in the bathroom.
First, focus on a seed. Then, make changes to the environment that grow the
plant until it reaches the reproduction life stage.

\promptgap
\reflectionlabel To grow the orange plant, I first need a seed. \hl{The task says seeds are in the bathroom}, so I teleport there.\\
\actionlabel teleport to bathroom\\
\observationlabel You teleport to the bathroom.

\promptgap
\reflectionlabel I look around the bathroom to find the seeds and necessary
items for planting.\\
\actionlabel look around\\
\observationlabel This room is called the bathroom. In it, you see:\\
\hspace*{1em}the agent\\
\hspace*{1em}a substance called air\\
\hspace*{1em}a bathtub, which is turned off. In the bathtub is: nothing.\\
\hspace*{1em}a glass cup (containing nothing)\\
\hspace*{1em}a painting\\
\hspace*{1em}a seed jar (containing a orange seed, a orange seed, a orange seed,
a orange seed, a orange seed)\\
\hspace*{1em}a sink, which is turned off. In the sink is: nothing.\\
\hspace*{1em}a toilet. In the toilet is: A drain, which is closed, a substance
called water.\\
You also see:\\
\hspace*{1em}A door to the kitchen (that is open)

\promptgap
\reflectionlabel I see a seed jar here; I pick it up so I can plant a seed.\\
\actionlabel pick up seed jar\\
\observationlabel You move the seed jar to the inventory.

\promptgap
\reflectionlabel I move an orange seed from the jar into the bathtub \hl{so it has
a medium to grow in}\\
\actionlabel move orange seed in seed jar to bathtub\\
\observationlabel You move the orange seed to the bathtub.

\promptgap
\reflectionlabel The task says to focus on the seed, so I focus on the orange
seed now that it is planted.\\
\actionlabel focus on orange seed in bathtub\\
\observationlabel You focus on the orange seed.

\promptgap
\reflectionlabel Seeds need water to grow, so I turn on the sink to get a water
source.\\
\actionlabel activate sink\\
\observationlabel The sink is now activated.

\promptgap
\reflectionlabel I move the glass cup to the running sink to fill it with
water.\\
\actionlabel move glass cup to sink\\
\observationlabel You move the glass cup to the sink.

\promptgap
\reflectionlabel I pour the water from the glass cup into the bathtub to water
the seed.\\
\actionlabel pour glass cup into bathtub\\
\observationlabel You pour the contents of the glass cup into the bathtub.

\promptgap
\reflectionlabel I turn the sink off now that the glass cup is emptied into the
bathtub.\\
\actionlabel deactivate sink\\
\observationlabel The sink is now deactivated.

\promptgap
\reflectionlabel Growth takes time, so I wait to let the plant develop.\\
\actionlabel wait1\\
\observationlabel You decide to wait for 1 iterations.

\promptgap
\textit{[... more waiting ...]}

\promptgap
\reflectionlabel I keep waiting to give the plant more time to grow.\\
\actionlabel wait1\\
\observationlabel You decide to wait for 1 iterations.

\promptgap
\reflectionlabel Plants need repeated watering as they grow, so I turn the sink
on again.\\
\actionlabel activate sink\\
\observationlabel The sink is now activated.

\promptgap
\reflectionlabel I refill the glass cup at the sink.\\
\actionlabel move glass cup to sink\\
\observationlabel You move the glass cup to the sink.

\promptgap
\reflectionlabel I water the growing plant again by pouring the glass cup into
the bathtub.\\
\actionlabel pour glass cup into bathtub\\
\observationlabel You pour the contents of the glass cup into the bathtub.

\promptgap
\reflectionlabel I turn the sink off after watering.\\
\actionlabel deactivate sink\\
\observationlabel The sink is now deactivated.

\promptgap
\reflectionlabel I wait to let the plant keep growing toward the reproduction
stage.\\
\actionlabel wait1\\
\observationlabel You decide to wait for 1 iterations.

\promptgap
\textit{[... more waiting ...]}

\promptgap
\reflectionlabel \hl{The plant may need more nutrients or sunlight to progress
further.} I should check if there are any additional resources in other rooms,
such as the greenhouse or outside.\\
\actionlabel teleport to greenhouse\\
\observationlabel You teleport to the greenhouse.

\promptgap
\reflectionlabel I look around the greenhouse to find resources that might help
the orange plant grow, such as nutrients or sunlight.\\
\actionlabel look around\\
\observationlabel This room is called the greenhouse. In it, you see:\\
\hspace*{1em}the agent\\
\hspace*{1em}a substance called air\\
\hspace*{1em}a adult bee\\
\hspace*{1em}a bee hive. The bee hive door is open. In the bee hive is:
nothing.\\
\hspace*{1em}a flower pot 1 (containing soil, a substance called water)\\
\hspace*{1em}a flower pot 2 (containing soil, a substance called water)\\
\hspace*{1em}a flower pot 3 (containing soil, a substance called water)\\
\hspace*{1em}a jug (containing nothing)\\
\hspace*{1em}a sink, which is turned off. In the sink is: nothing.\\
You also see:\\
\hspace*{1em}A door to the hallway (that is open)\\
\hspace*{1em}A door to the outside (that is open)

\promptgap
\reflectionlabel \hl{The greenhouse has flower pots with soil and water, but I
already have a water source.} I can use the jug here to collect more water if
needed, but I should check outside for sunlight, which is essential for plant
growth.\\
\actionlabel teleport to outside\\
\observationlabel You teleport to the outside.

\promptgap
\reflectionlabel I am now outside, where the plant can receive sunlight. I
should ensure the orange plant in the bathtub has access to sunlight, so I will
return to the bathroom and check on its progress.\\
\actionlabel teleport to bathroom\\
\observationlabel You teleport to the bathroom.

\promptgap
\reflectionlabel \hl{I return to the bathroom to check on the orange plant in the
bathtub and ensure it is receiving adequate sunlight.} I will continue to water
it and monitor its growth until it reaches the reproduction stage.\\
\actionlabel focus on orange seed in bathtub\\
\observationlabel You focus on the orange seed.

\promptgap
\reflectionlabel \hl{The orange seed has likely begun to germinate and grow into a
seedling. I will continue to water it regularly and ensure it receives sunlight
by keeping the bathroom window open if possible.} I will wait for more
iterations to allow the plant to progress toward the reproduction stage.\\
\actionlabel wait1\\
\observationlabel You decide to wait for 1 iterations.

\promptgap
\textit{[... more waiting ...]}

\promptgap
\textbf{Outcome:} Failed\\
\textbf{Reward:} 39\\
\textbf{Environment steps:} 30\\
\textbf{WALL-E rejections:} 0
\end{PromptBlock}

\promptheading{BB-WM}

\begin{PromptBlock}
Task Description:\\
\tasklabel grow a orange plant from seed. Seeds can be found in the bathroom.
First, focus on a seed. Then, make changes to the environment that grow the
plant until it reaches the reproduction life stage.

\promptgap
\reflectionlabel To grow the orange plant, I first need a seed. The task says
seeds are in the bathroom. \hl{Before I start exploring, I'll consult the world
model to confirm where the orange seed is most likely to be.}\\
\actionlabel query where is orange seed\\
\observationlabel orange seed is most likely in: bathroom (0.17), bedroom
(0.17), greenhouse (0.17), kitchen (0.17), living room (0.17), workshop
(0.17).

\promptgap
\reflectionlabel The world model indicates that the orange seed is equally
likely to be in several rooms, including the bathroom. \hl{Since the task specifies
that seeds can be found in the bathroom, I should start by teleporting there to
search for the seed.}\\
\actionlabel teleport to bathroom\\
\observationlabel You teleport to the bathroom.

\promptgap
\reflectionlabel I look around the bathroom to locate the orange seed. If it's
not here, I will need to check other rooms based on the world model's
probabilities.\\
\actionlabel look around\\
\observationlabel This room is called the bathroom. In it, you see:\\
\hspace*{1em}the agent\\
\hspace*{1em}a substance called air\\
\hspace*{1em}a bathtub, which is turned off. In the bathtub is: nothing.\\
\hspace*{1em}a glass cup (containing nothing)\\
\hspace*{1em}a painting\\
\hspace*{1em}a seed jar (containing a orange seed, a orange seed, a orange
seed, a orange seed, a orange seed)\\
\hspace*{1em}a sink, which is turned off. In the sink is: nothing.\\
\hspace*{1em}a toilet. In the toilet is: A drain, which is closed, a substance
called water.\\
You also see:\\
\hspace*{1em}A door to the kitchen (that is open)

\promptgap
\reflectionlabel I found a seed jar containing orange seeds in the bathroom. I
will pick it up to proceed with planting the seed.\\
\actionlabel pick up seed jar\\
\observationlabel You move the seed jar to the inventory.

\promptgap
\reflectionlabel I need to select an orange seed from the jar and plant it in
a suitable container. I will move one of the seeds to a flower pot, \hl{but first,
I need to check if there are flower pots available in the current room or
elsewhere. I'll consult the world model to find a flower pot.}\\
\actionlabel query where is flower pot\\
\observationlabel flower pot is most likely in: greenhouse (1.00).

\promptgap
\reflectionlabel The world model indicates that the flower pot is most likely
in the greenhouse. I will teleport there to plant the orange seed in a suitable
container.\\
\actionlabel teleport to greenhouse\\
\observationlabel You teleport to the greenhouse.

\promptgap
\reflectionlabel I am now in the greenhouse, where the flower pot is likely
located. I will look around to find a flower pot to plant the orange seed.\\
\actionlabel look around\\
\observationlabel This room is called the greenhouse. In it, you see:\\
\hspace*{1em}the agent\\
\hspace*{1em}a substance called air\\
\hspace*{1em}a adult bee\\
\hspace*{1em}a adult bee\\
\hspace*{1em}a adult bee\\
\hspace*{1em}a adult bee\\
\hspace*{1em}a bee hive. The bee hive door is open. In the bee hive is:
nothing.\\
\hspace*{1em}a flower pot 1 (containing soil, a substance called water)\\
\hspace*{1em}a flower pot 2 (containing soil, a substance called water)\\
\hspace*{1em}a flower pot 3 (containing soil, a substance called water)\\
\hspace*{1em}a jug (containing nothing)\\
\hspace*{1em}a sink, which is turned off. In the sink is: nothing.\\
You also see:\\
\hspace*{1em}A door to the hallway (that is open)\\
\hspace*{1em}A door to the outside (that is open)

\promptgap
\reflectionlabel I see three flower pots in the greenhouse, all containing
soil and water. I will move an orange seed from the seed jar in my inventory
to one of the flower pots, starting with flower pot 1.\\
\actionlabel move orange seed in seed jar to flower pot 1\\
\observationlabel You move the orange seed to the flower pot 1.

\promptgap
\reflectionlabel I have successfully planted the orange seed in flower pot 1.
Next, I need to focus on the seed to begin the growth process.\\
\actionlabel focus on orange seed in flower pot 1\\
\observationlabel You focus on the orange seed.

\promptgap
\reflectionlabel \hl{The seed needs water to grow.} I will activate the sink to get
a water source and use the jug to water the seed.\\
\actionlabel activate sink\\
\observationlabel The sink is now activated.

\promptgap
\reflectionlabel I will move the jug to the sink to fill it with water.\\
\actionlabel move jug to sink\\
\observationlabel You move the jug to the sink.

\promptgap
\reflectionlabel The jug is now at the sink. I will pour water from the jug
into flower pot 1 to water the seed.\\
\actionlabel pour jug into flower pot 1\\
\observationlabel You pour the contents of the jug into the flower pot 1.

\promptgap
\reflectionlabel I have watered the seed. Now I need to wait for the plant to
grow toward the reproduction stage. I will deactivate the sink and then wait
for a few steps.\\
\actionlabel deactivate sink\\
\observationlabel The sink is now deactivated.

\promptgap
\reflectionlabel \hl{The plant needs time to grow, so I will start waiting to
allow it to progress through its life stages.}\\
\actionlabel wait1\\
\observationlabel You decide to wait for 1 iterations.

\promptgap
\reflectionlabel I will continue waiting to give the plant more time to grow
toward the reproduction stage.\\
\actionlabel wait1\\
\observationlabel You decide to wait for 1 iterations.

\promptgap
\textit{[... more waiting ...]}

\promptgap
\reflectionlabel I have waited several steps for the plant to grow. I should
now check if the plant has reached the reproduction stage or if further
actions, such as watering again, are needed. I will first look around to
observe the current state of the flower pot and the plant.\\
\actionlabel look around\\
\observationlabel This room is called the greenhouse. In it, you see:\\
\hspace*{1em}the agent\\
\hspace*{1em}a substance called air\\
\hspace*{1em}a adult bee\\
\hspace*{1em}a adult bee\\
\hspace*{1em}a bee hive. The bee hive door is open. In the bee hive is:
nothing.\\
\hspace*{1em}a flower pot 1 (containing a orange tree in the seedling stage,
soil, a substance called water)\\
\hspace*{1em}a flower pot 2 (containing soil, a substance called water)\\
\hspace*{1em}a flower pot 3 (containing soil, a substance called water)\\
\hspace*{1em}a sink, which is turned off. In the sink is: a jug (containing a
substance called water).\\
You also see:\\
\hspace*{1em}A door to the hallway (that is open)\\
\hspace*{1em}A door to the outside (that is open)

\promptgap
\reflectionlabel The orange plant in flower pot 1 is now in the seedling
stage. It still needs time and care to progress to the reproduction stage. I
will water it again by activating the sink, refilling the jug, and pouring
water into the flower pot.\\
\actionlabel activate sink\\
\observationlabel The sink is now activated.

\promptgap
\reflectionlabel I will move the jug to the sink to refill it with water.\\
\actionlabel move jug to sink\\
\observationlabel You move the jug to the sink.

\promptgap
\reflectionlabel The jug is now at the sink. I will pour water from the jug
into flower pot 1 to water the seedling again.\\
\actionlabel pour jug into flower pot 1\\
\observationlabel You pour the contents of the jug into the flower pot 1.

\promptgap
\reflectionlabel I have watered the seedling again. Now I need to deactivate
the sink and continue waiting for the plant to progress toward the reproduction
stage.\\
\actionlabel deactivate sink\\
\observationlabel The sink is now deactivated.

\promptgap
\reflectionlabel The plant is still in the seedling stage and requires more
time to grow. I will wait for a few more steps to allow it to progress
further.\\
\actionlabel wait1\\
\observationlabel You decide to wait for 1 iterations.

\promptgap
\textit{[... more waiting ...]}

\promptgap
\textbf{Outcome:} Success\\
\textbf{Reward:} 100\\
\textbf{Environment steps:} 25\\
\textbf{Belief query steps:} 2\\
\textbf{WALL-E rejections:} 0
\end{PromptBlock}


\newpage

\end{document}